\pdfoutput=1

\documentclass[11pt]{article}
\usepackage[final]{acl}

\usepackage{times}
\usepackage{latexsym}

\usepackage[T1]{fontenc}

\usepackage[utf8]{inputenc}

\usepackage{microtype}

\usepackage{inconsolata}

\usepackage{graphicx}

\usepackage{amsmath}
\usepackage{amssymb}
\usepackage{enumitem}
\usepackage{booktabs}
\usepackage{multirow}
\usepackage{tcolorbox}
\usepackage{xspace}
\usepackage{bm}

\newcommand{\basequ}{SET?\xspace}
\newcommand{\baseex}{SET!\xspace}

\def\argmax{\mbox{argmax}}
\def\dnrm#1{\mbox{$_{\hbox{\scriptsize #1}}$}}
\def\numberlanguagesaccurate{1665\xspace}
\def\numberlanguagestotal{1832\xspace}
\def\numberlanguagediff{167\xspace}
\def\numlangslidcommunity{300\xspace}
\def\modelname{\mbox{GlotLID-M}\xspace}
\def\corpusname{\mbox{GlotLID-C}\xspace}
\def\genericname{\mbox{GlotLID}\xspace}
\def\macrolanguage{macrolanguage\xspace}
\def\macrolanguages{macrolanguages\xspace}
\def\flores{FLORES\xspace} 
\def\udhr{UDHR\xspace}

\def\ft176{FT176\xspace}
\def\edin{OpenLID\xspace}
\def\nllb{NLLB\xspace}

\def\tablebreak{\rule{0pt}{6pt} \nolinebreak\hspace{\fill}\linebreak
}

\def\figref#1{Figure~\ref{fig:#1}}
\def\figlabel#1{\label{fig:#1}\label{p:#1}}

\def\tabref#1{Table~\ref{tab:#1}}

\def\tablabel#1{\label{tab:#1}\label{p:#1}}
\def\eqref#1{Eq.~\ref{eqn:#1}}

\def\seclabel#1{\label{sec:#1}\label{p:#1}}
\def\secref#1{\S\ref{sec:#1}}

\title{\genericname: Language Identification for
  Low-Resource Languages}

\author{Amir Hossein Kargaran$^{\clubsuit}$, Ayyoob Imani$^{\clubsuit}$, François Yvon$^{\spadesuit}$ and Hinrich Schütze$^{\clubsuit}$ \\
        $^\clubsuit$LMU Munich \& Munich Center for Machine Learning, Munich, Germany \\
        $^\spadesuit$Sorbonne Université \& CNRS, ISIR, Paris, France \protect \\
        \texttt{amir@cis.lmu.de}
}

\begin{document}
\maketitle
\begin{abstract}
Several recent papers have published good solutions for
language identification (LID) for about \numlangslidcommunity high-resource
and medium-resource languages. However, there is no LID
available that (i) covers a wide range of low-resource languages, (ii)
is rigorously evaluated and reliable and (iii) efficient and
easy to use.  Here, we publish \modelname, an LID model that
satisfies the desiderata of wide coverage, reliability and
efficiency.  It identifies \numberlanguagesaccurate
languages, a large increase in coverage compared to prior work. In our experiments, \modelname
outperforms
four baselines (CLD3, \ft176, \edin and \nllb) when
balancing F1 and false positive rate (FPR).
We analyze
the unique challenges that
low-resource LID poses: incorrect corpus metadata, leakage
from high-resource languages, difficulty separating closely
related languages, handling of \macrolanguage vs varieties
and in general noisy data.  We hope that
integrating \modelname into dataset creation pipelines will
improve quality and enhance accessibility of NLP technology
for low-resource languages and cultures. \modelname model (including future versions), code, and list of data sources are available at: \url{https://github.com/cisnlp/GlotLID}. 

\begin{tcolorbox}
The evaluation results in this paper are based on \modelname v1.0. \modelname{} undergoes regular updates, and previous versions are also maintained. As of April 2024, \modelname v3.0, which includes over 2000 labels, has been released.
\end{tcolorbox}
\end{abstract}

\section{Introduction}
The NLP community should 
create technology that covers as many languages as possible,
not only medium-resource and high-resource languages.  This
goal can only be achieved if corpora for low-resource
languages are available.  Web-mined datasets -- including
CC100~\citep{wenzek-etal-2020-ccnet}, mC4~\citep{xue-etal-2021-mt5}
and OSCAR~\citep{AbadjiOrtizSuarezRomaryetal.2021,
OrtizSuarezSagotRomary2019} -- have made important
contributions to low-resource NLP. In
particular, they lay the ground for multilingual neural
models like XLM-R~\citep{conneau-etal-2020-unsupervised},
mT5~\citep{xue-etal-2021-mt5} and
Glot500~\citep{imanigooghari-etal-2023-glot500}. However, existing
web-mined datasets have systematic quality issues
\citep{kreutzer-etal-2022-quality} and insufficient coverage of
low-resource languages.

Low-quality datasets cause poor performance for downstream applications. They can
also give rise to a misleading perception of progress when 
coverage of a low-resource language is claimed based on
noisy data.  NLP for low-resource languages
requires high-quality datasets and high-quality datasets
require high-quality LID (language identification).  For this reason, high-quality LID
for low-resource languages is paramount.  To address this
need, in this paper we present \modelname, a high-quality
LID that covers
\numberlanguagesaccurate languages.
We use ISO
639-3 to individuate languages.

When expanding the scope of LID  from
a few hundred
to \numberlanguagesaccurate
languages, the problem of \emph{granularity} becomes
severe. In real-world settings, LID needs to
support both \macrolanguages and their varieties; it also needs to
be
robust against out-of-model cousins \citep{caswell-etal-2020-language, kreutzer-etal-2022-quality}.
We pay particular attention to this issue.

While low-resource is our main focus, \citet{blevins-zettlemoyer-2022-language} point out
that low-quality LID also affects high-resource corpora
through contamination, resulting in claims of successful
crosslingual transfer that are due to unrecognized
coverage of low-resource languages. We also address this
issue, e.g., we improve English F1 on  the ``Universal Declaration of Human Rights'' corpus (\udhr) to .85
compared to .43 for \edin.

\textbf{Contributions.}
(i) We curate \corpusname, a comprehensive dataset
covering \numberlanguagesaccurate languages, most of
them low-resource, from a diverse set of domains. 
(ii) We train \modelname
on \corpusname, 
an open-source LID covering
these \numberlanguagesaccurate languages.
(iii) In our experiments, \modelname outperforms several baselines
by more than 12\% absolute F1
on \udhr, which we take as the best benchmark for our focus
on low-resource languages.
(iv) When balancing F1 and false positive rate (FPR), \modelname
also outperforms baselines on \flores-200, which is dominated
by high-/medium-resource languages.

\section{Requirements for low-resource LID}
\seclabel{reqs}

\textbf{Main use case: Corpus creation.}
Corpus creation and cleaning is the main use case for our
low-resource LID because we want to address the need for
high-quality corpora for low-resource languages.  Line-by-line
LID filtering is an effective method for achieving high
corpus quality. Reliable LID can eliminate various types
of noise (see \citep{caswell-etal-2020-language,kreutzer-etal-2022-quality})
-- including data from other languages
and non-linguistic data -- that is frequent, especially in
web-crawled content. By adjusting the confidence threshold, users
will have control over the level of quality of
the corpora they create.

\textbf{Broad coverage of languages, minimize out-of-model cousin errors.} 
We strive for as broad a coverage as is possible given
available datasets. This has two benefits. First,
it reduces ``out-of-model cousin''
errors \citep{caswell-etal-2020-language, kreutzer-etal-2022-quality},
i.e., it reduces the risk that a language not covered is misclassified
as a closely related covered language.
Second, having LIDs that discriminate many low-resource languages is a pre-requisite
for developing NLP technologies for the largest possible number of languages.
Yet many existing LIDs only cover a few hundred
languages. In this study, we therefore focus on LIDs
having a broad
coverage, excluding
CLD2~\citep{mccandless2010accuracy},
Equilid~\citep{jurgens-etal-2017-incorporating}, 
Langdetect~\citep{shuyo2010language} and
langid.py~\citep{lui-baldwin-2012-langid}.
These LIDs cover less than 100~languages or are
outperformed by the models we compare with.

\textbf{Open-source.}
LIDs should be open-source to encourage
open collaboration and conform to best research practices.
Some LIDs that meet our other requirements
are not open-source, e.g., those published by
\citet{caswell-etal-2020-language, bapna2022building, kudugunta2023madlad}.
CLD3~\citep{botha-etal-2017-natural, salcianu2018compact}
is freely available, but its training code is not open-source.

\textbf{Ease of use.}
LIDs should be
easily deployable across platforms and programming environments
without having to worry about dependencies, compatibility
and lack of maintenance.

Because of this ease-of-use requirement, we do not consider
whatlang \citep{brown-2014-non, la-strings} nor idNet
\citep{dunn2020mapping}, two broad-coverage
LIDs that meet many other requirements, but are hard
to use in many practical scenarios due to software issues and lack of maintenance.

\textbf{Uncertainty assessment.} In our use cases, we
would like to rely on uncertainty measures to 
distinguish cases where the highest-probability language is
certain from those where it is not. This would allow us to
choose a level of confidence for the resulting corpus. For
example, we may want to retain only sentences
identified with a high confidence (say, 70\%). This is essential to
produce high-quality low-resource corpora.

Because of this requirement, we do not consider 
Franc~\citep{franc} as a baseline.
While it has many desirable properties, it generally does
not provide well-calibrated probabilities. It
usually returns several classes, giving 1.0 to the top
class and values close to 1.0 to several others.

\textbf{Efficiency.} LID is easy to run in parallel, but we
still need an efficient solution to make it applicable
to large corpora, not least for ecological reasons. 

Lack of efficiency is the reason why we do not use
AfroLID~\citep{adebara-etal-2022-afrolid} as a baseline,
despite its excellent coverage of
African languages.\footnote{It has no coverage of other low-resource languages.} 
AfroLID
is a transformer
architecture and
less efficient than its competitors.

\textbf{Granularity flexibility.} When scaling LID from a
few hundred languages to more than 1500, it is hardly practical
to restrict the set of labels to a single
level of the language hierarchy (e.g., using resources like
\href{https://iso639-3.sil.org}{\path{iso639-3.sil.org}}). This is  due to the complexity of defining
and delimiting languages, including the
coexistence of \macrolanguages and their varieties. In many
cases, we  want to keep both the \macrolanguage and the
varieties in our label set because the varieties we have data
for are
important languages in their own right. But for other
varieties, we do not have variety-labeled data, so the only
way to include them is through
the \macrolanguage. For example, \flores-200~\citep{nllbteam2022language} covers the \macrolanguage
aka (Akan) and its variety twi (Twi), but not its variety
fat (Fanti).
Keeping both aka and twi gives flexibility to 
LID users: they can either differentiate aka and twi
or they can
consolidate the two labels to the single label aka,
depending on what makes more sense in their setting.

\section{Dataset curation}
\seclabel{glotliddataset}
We now describe \corpusname, a corpus for  LID training that covers 
\numberlanguagestotal languages.

\textbf{Source selection.}
We choose sources that we deem trustworthy (i.e., high
chance of correct language label).
To address the domain sensitivity of
LID  and broaden language coverage, we curate
a diverse set of text domains.

We review sources referenced
by \citet{imanigooghari-etal-2023-glot500, burchell-etal-2023-open, blaschke-etal-2023-survey, adebara-etal-2022-afrolid, adebara-abdul-mageed-2022-towards}.
In each case, we consider the collection methodology,
selecting sources whose language labels are trustworthy.
We generally do not use web-crawled sources
to avoid the associated problems
\citep{kreutzer-etal-2022-quality}. 
Most selected sources are derived from Wikipedia, religious
texts, collaborative translations, storybooks, and news
sites. This gives us a coverage
of \numberlanguagestotal languages, more than any other
public LID. For a list of data sources, see \secref{datasources}.

\textbf{Preprocessing.} We ensure that each sentence is written in the correct script, based on the writing system databases of \citet{kargaran2023glotscript} and \citet{van-esch-etal-2022-writing}. We use the GlotScript~\citep{kargaran2023glotscript} Python library to determine scripts. We also eliminate duplicate sentences.

\textbf{Statistics.}
Our final corpus, \corpusname, comprises 289 million
sentences (i.e., lines of data) totaling 40GB and spans
\numberlanguagestotal languages (identified by their ISO
639-3 code). 1677
languages have more than 1000 sentences.
Refer to \secref{appendix_glot_performance} for the total number of sentences per language.

\textbf{Train/test split.}
We designate 85\% of the data
as \textbf{\corpusname train}.
Let $n_l$ be the number of sentences from language $l$ 
in the remaining 15\%. 
Then we sample $\min(1000,n_l)$ sentences from it.
We refer to the resulting dataset
as \textbf{\corpusname test}.

\textbf{Contamination.} To make sure our evaluation data (especially \udhr, refer to \secref{evaluation_data})
do not overlap with our sources, we compute contamination of \udhr in \corpusname train.

We count a \udhr test sentence as occurring in the training set if all of its word four-grams occur in one sentence of \corpusname. 
Most of these contaminations are due to two resources: Wikipedia and
Tatoeba~\citep{tatoeba}. \corpusname train shares 374 languages with \udhr.

For 292 languages, we find that none of the \udhr test sentences occurs in the training data. For 57 languages, less than 10\% of \udhr test sentences occur in the training data. The remaining 25~languages with a contamination rate over 10\% are all high/medium resource languages.

In our experiments, we decided against removing any sentences from \corpusname, as there is little contamination of \udhr for low-resource languages. We follow here most prior work which has the problem of contamination of \udhr for high-resource languages. We will however remove from \corpusname train the sentences causing contamination as part of our next release.

\subsection{List of data sources}\seclabel{datasources}

\begin{itemize}[noitemsep]
    \item \textbf{Wikipedia articles:} WiLI-2018~\citep{thoma2018wili}, Leipzig corpora-wiki~\citep{goldhahn-etal-2012-building}, Wikipedia dumps~\citep{wikipedia-dumps}
    \item \textbf{News:} BBC News~\citep{hasan-etal-2021-xl}, Global Voices \citep{tiedemann-2012-parallel}, Leipzig corpora-news~\citep{goldhahn-etal-2012-building}, SETIMES~\citep{tiedemann-2012-parallel}
    \item \textbf{Translation:} NLLB Seed~\citep{nllbteam2022language}
    \item \textbf{Religious:} PBC~\citep{mayer-cysouw-2014-creating}, Jehovah's Witnesses~\citep{jw}, 1000Langs~\citep{1000langs-repo}
    \item \textbf{Crowdsourcing:} Tatoeba~\citep{tatoeba}
    \item \textbf{Multiple domain:}  MT-560 \citep{gowda-etal-2021-many,tiedemann-2012-parallel, burchell-etal-2023-open, post-etal-2012-constructing,ziemski-etal-2016-united,rozis-skadins-2017-tilde,kunchukuttan-etal-2018-iit,qi-etal-2018-pre,zhang-etal-2020-improving,bojar-etal-2013-findings,bojar-etal-2014-findings,bojar-etal-2015-findings,bojar-etal-2016-findings,bojar-etal-2017-findings,bojar-etal-2018-findings,barrault-etal-2019-findings,barrault-etal-2020-findings}, LTI~\citep{brown2012finding}, Arabic~\citep{zahir2022iadd,alsarsour-etal-2018-dart,abu-kwaik-etal-2018-shami,medhaffar-etal-2017-sentiment,meftouh-etal-2015-machine,zaidan-callison-burch-2011-arabic, el-haj-etal-2018-arabic, bouamor-etal-2019-madar},  Persian~\citep{pilevar2011tep, kashefi2018mizan}, Turkic~\citep{mirzakhalov-etal-2021-large}, Bhojpuri~\citep{ojha2019english}, Cantonese~\citep{luke2015hong}, Guaran{\'i}~\citep{gongora-etal-2022-use}, Manipuri~\citep{huidrom-etal-2021-em}
    \item \textbf{Government domain:} Autshumato~\citep{groenewald2010processing}
\end{itemize}

We also introduce additional data sources suited for LID, although they are not included in the training of the version of \modelname discussed in the paper. Specifically, GlotSparse\footnote{\url{https://github.com/cisnlp/GlotSparse}} (collection of news websites in low-resource lan-
guages) and GlotStoryBook\footnote{\url{https://github.com/cisnlp/GlotStoryBook}} (collection of children storybooks) are two corpora compiled as part of this project to include more languages and domains for LID. These data sources will be used in future versions of \modelname. These data sources include:

\begin{itemize}[noitemsep]
    \item \textbf{Crowdsourcing:}
    CommonVoice v11~\citep{ardila-etal-2020-common}
    \item \textbf{Web:} Wanca 2016~\citep{jauhiainen2019wanca}
    \item \textbf{News:} GlotSparse, MasakhaNEWS~\citep{Adelani2023MasakhaNEWS}, Goud.ma~\citep{issam2022goudma}, AI4D~\citep{siminyu2021ai4d}, Radio Ramogi~\citep{luo-news-dataset}, smugri~\citep{yankovskaya-etal-2023-machine}, finno-ugric~\citep{yankovskaya-etal-2023-machine}
    \item \textbf{Trasnlation:} AfriQA~\citep{ogundepo2023afriqa}, smugri-flores~\citep{yankovskaya-etal-2023-machine}, GlotStoryBook
    \item \textbf{Multiple domain:} Universal Dependencies v2.12~\citep{nivre-etal-2020-universal}, Abkhaz National Corpus~\citep{meurer-2018-abkhaz}
    \item \textbf{Lyrics:} lyricstranslate~\citep{lyricstranslate}
    \item \textbf{Government domain:} Vuk'uzenzele~\citep{lastrucci-etal-2023-preparing}
\end{itemize}

\section{\modelname}

We select
FastText~\citep{joulin-etal-2017-bag}
as the architecture for \modelname, because it satisfies all
requirements
outlined in \secref{reqs} as we will explain now.

We train our FastText model \modelname on \corpusname train 
with \numberlanguagestotal languages. FastText can easily handle the large
number of languages in the
corpus. Because of this \textbf{broad coverage}, \textbf{out-of-model
cousin errors are reduced}. 
Although we restrict
the number of classes to \numberlanguagesaccurate for some
experiments (e.g., in \tabref{glotlidresults}),
\modelname's  classification always uses
all \numberlanguagestotal languages to mitigate
out-of-model cousin errors.
This satisfies the first
requirement from \secref{reqs}: \modelname is a useful tool
for corpora creation because it has a broad coverage of
languages that can occur in raw data.

FastText provides an \textbf{open-source} codebase for
training, which supports customization and
extension of \modelname.

FastText is \textbf{easy to use}: It offers a number of
language bindings, making it compatible with multiple
programming languages (including C++, Python, Java, Node.js, Rust, Ruby, R) and reducing
dependency, incompatibility and other software issues.

FastText
meets the requirement of \textbf{uncertainty assessment}
because it
provides confidence
scores that can serve as  thresholds to
effectively mitigate noise in the data.  
For the same reason, FastText also supports \textbf{granularity
flexibility}: we can accumulate probabilities over language varieties to get
a good estimate of the probability of the \macrolanguage.
To this end, we simply add to the \macrolanguage probability
the probabilities of its varieties. This way, the system can
return appropriate estimates at various levels of granularity.

As a professionally designed and implemented linear classifier,
FastText is \textbf{efficient}: it had the best throughput of the candidate solutions
we tested and can process large corpora with high speed.
As a linear model, FastText has the additional advantage of 
delivering explainable classification decisions.
FastText is a multinomial logistic classifier. The input
sentence
is represented as an average of  n-gram embeddings. 
This allows us to visualize
how much each n-gram contributed to the final prediction.
See
\citet{nllbteam2022language}, Fig.\ 8, for details.

Taking all these requirements together (and its good LID performance demonstrated in \secref{results} and acceptable calibration in \secref{calibration}), \modelname, based on FastText, is, in our opinion, an excellent tool for supporting our use case, the \textbf{creation of high-quality low-resource corpora}.

\section{Experimental setup}
We train \modelname on \corpusname train
using the hyperparameters
in \citep{nllbteam2022language, burchell-etal-2023-open} and
otherwise FastText defaults (see
\secref{hyperparameters}).
Following \citet{arivazhagan2019massively}, \citet{nllbteam2022language}
and \citet{burchell-etal-2023-open}, we perform up-sampling
for low resource languages.  Sentences from a language $l$
representing $p_l$ of the dataset are sampled
proportionally to $p_l^{\frac{1}{T}}$ where $T$ is the
temperature.  Following \citet{nllbteam2022language}
and \citet{burchell-etal-2023-open}, we set
$\frac{1}{T}=.3$.

\subsection{\modelname hyperparameters}\seclabel{hyperparameters}

We provide the hyperparameters used to train the \modelname in \tabref{tabhyperparameters}.

\setlength{\textfloatsep}{5pt}
\begin{table}[h]
\small
 \centering
\resizebox{0.99\linewidth}{!}{
\begin{tabular}{llc}
\toprule
{Argument} & {Description} & {Value} \\ \midrule
{-minCount} & {Minimal number of word occurrences} & {1000}  \\
{-minCountLabel} & {Minimal number of label occurrences} & {0} \\
{-wordNgrams} & {Max length of word ngram} & {1} \\
{-bucket} & {Number of buckets} & {10$^{6}$} \\
{-minn} & {Min length of char ngram} & {2} \\
{-maxn} & {Max length of char ngram} & {5} \\
{-loss} & {Loss function} & {softmax} \\
{-dim} & {Size of word vectors} & {256} \\
{-epoch} & {Number of epochs} & {2} \\
{-lr} & {Learning rate} & {.8} \\
\bottomrule
\end{tabular}
}
\caption{\modelname training hyperparameters}
\tablabel{tabhyperparameters}
\end{table}

\subsection{Evaluation data}\seclabel{evaluation_data}

We evaluate \modelname on \corpusname test,
\flores-200~\citep{nllbteam2022language} and
\udhr\footnote{\url{http://www.unicode.org/udhr/d/}} (Universal
Declaration of Human Rights).

While testing on  data unseen in training is standard in NLP,
the results have to be taken with a grain of salt because there is often
a domain mismatch in real-world applications of LID
\citep{caswell-etal-2020-language, dunn2020mapping}.
\flores-200 and \udhr 
address this concern: they are not part of our training set
(however, see discussion in \secref{glotliddataset})
and do not draw on our sources. Many other
benchmarks share sources like Wikipedia with us
\citep{thoma2018wili, haas-derczynski-2021-discriminating,
ahmadi-etal-2023-pali}. 
\flores-200 and \udhr  are also the benchmarks
with the broadest available language coverage.

\textbf{\flores-200}
is a collection of 842 articles
obtained from English-language Wikimedia projects. Each
sentence in the articles was translated into 
204 distinct language-script combinations, corresponding to
196 distinct languages,
and
human-verified. It
provides 997 sentences for development, 1012  for
dev-test and 992 for  test. \flores-200 test is not
publicly available.
Following prior work,
we use
dev-test as our \flores test set.

The level of granularity
across language (sub)families varies in \flores; e.g.,
it
includes nine varieties of Arabic.
On the other hand, some languages (e.g., 
est:Estonian) are only
available as \macrolanguage.
In some cases, \flores includes
both a \macrolanguage and varieties, e.g.,
 aka
(Akan) and its variety twi (Twi), and zho (Chinese) and its
variety yue (Yue Chinese).
Although some issues have been reported
(see \secref{flores-issues}) with \flores, we do
not have the resources to investigate them, so we
use it as is.

\textbf{\udhr}
consists of more than 500 translations of the
``Universal Declaration of Human Rights''.
419 translations available from the ``UDHR in Unicode'' project have a
iso-639-3 code that is not ``und'' (undetermined). 
We discard short
sentences (e.g., consisting
of just an article number or the single English word ‘missing’) by discarding the 35\% shortest
sentences for each language.

In some cases (e.g., Zulu and Quechua), UDHR
contains both a macrolanguage and one of its varieties. We have also seen some issues in \udhr (see \secref{udhr-issues}), but we have not extensively investigated these potential problems.

\subsection{Baselines}\seclabel{baselines}
Our baselines are \ft176,\footnote{\url{https://fasttext.cc/docs/en/language-identification.html}} CLD3, \nllb~\citep{nllbteam2022language} and \edin~\citep{burchell-etal-2023-open}. The first two were used for filtering
the resources OSCAR and
mC4~\citep{kreutzer-etal-2022-quality}.

\textbf{CLD3.} CLD3 uses an n-gram (1$\leq$n$\leq$3)
based neural network model. CLD3 sometimes deviates from established
metadata conventions. For example, ISO-639-1 ku refers
to kur (Kurdish), but in CLD3 ku refers to its variety kmr (Northern Kurdish). It refers to Hebrew as iw, but the ISO code for Hebrew has changed to he and heb.

\textbf{\ft176.}
\ft176 
 is a FastText model
that uses
Wikipedia (WP) codes  as labels.
The documentation of language metadata is sometimes unclear;
e.g.,
\ft176 refers to Alemannic German as als although
ISO-639-3 als is Tosk Albanian.
It refers to the Malay \macrolanguage as ms, but unlike
ISO-639-3, this does not include ind (Indonesian).

\textbf{\nllb and \edin.} \nllb and \edin are
FastText
models. Their language label sets are mostly taken from
\flores, so granularity and coverage are similar to \flores.

\begin{figure*}
\centering
\begin{minipage}[t]{\textwidth}
\small
\textbf{Decision rule}

\smallskip

Given an LID classifier $m$, a base set $B$ of languages and
a threshold $\theta$, we assign label $\phi(s,m,B,\theta)$ to
sentence $s$ as follows:
\begin{equation*}
  \phi(s,m,B,\theta)=\begin{cases}
      \text{undetermined} & \text{if $\max_{l \in B}P_{m}(l|s)<\theta$}\\
 \argmax_{l \in B}  P_{m}(l|s)
           & \text{otherwise}
            \end{cases}
            \end{equation*}
           We distinguish two scenarios: \baseex and \basequ.

\smallskip

In scenario \baseex, the set of languages covered
      by the evaluation benchmark is
      known. We restrict a model's predictions to those languages that
      occur in the benchmark. This means that $B$ is a
      (proper or improper, see table captions for details)
      subset of the languages occurring in the benchmark.

\smallskip

In scenario \basequ, the set of languages covered by the
      evaluation benchmark is not known. We do not restrict
      a model $m$'s
      predictions: the model considers the entire set of
      languages it was trained on. This means that $B$ is
      the set of languages that $m$ was trained on.

\smallskip

\end{minipage}
\caption{\figlabel{decisionrule}Decision rule for assigning classes (i.e.,
      languages) in language identification}
\end{figure*}

\textbf{Language metadata matching.}
Matching the metadata
of the models to the metadata of the  benchmarks
(\flores, \udhr, \corpusname) is  not easy.
First, models do not consistently adhere to standard language
codes.
In addition, differences in granularity require
matching rules. For example, if a benchmark only covers
a \macrolanguage and none of its varieties, then
we consolidate classification decisions for the \macrolanguage and
its variations into the \macrolanguage label.

\textbf{Confidence thresholds.} For CLD3, we use
.5 and .7,
the two preset thresholds in Google's CLD3 repository.
For the other three baselines and \modelname, we also use
.5, but we use .3 as the second threshold value because .7
severely reduces the number of positive predictions for the
FastText models, resulting in low F1.

Prior work has not systematically
investigated the effect of confidence thresholding. However, it is of key
importance for our
use case of creating high-quality corpora for low-resource
languages. See \secref{result-scenario}
and \secref{results} for discussion of this point.

\subsection{Decision rule}\seclabel{result-scenario}
\figref{decisionrule} defines our decision rule.

\textbf{\baseex scenario.} When comparing LIDs $m_1$
and $m_2$ (trained on the set of languages $M_1$ and $M_2$)
on a benchmark $T$ (supporting the set of languages $B(T)$),
many evaluations  create a  subset
$M_1 \cap M_2 \cap B(T)$ and remove all sentences in the
benchmark that are labeled with languages outside of
$M_1 \cap M_2 \cap B(T)$.
\baseex evaluation
replicates this standard way of evaluating LIDs.

\textbf{\basequ scenario.} 
We believe that the \baseex scenario makes the LID task
unrealistically easy: a portion of the
data that could give rise to false positives
(data   not in $M_1 \cap M_2 \cap B(T)$)
is removed.
It is particularly unrealistic for our low-resource
scenario. Instead of hundreds of languages that are not
supported by all models, we have
more than a thousand. \emph{We therefore run
evaluations on the data for all languages} -- not just for
$M_1 \cap M_2 \cap B(T)$. That
is, we run evaluations on the entire benchmark $T$, not on
the subset in $M_1 \cap M_2 \cap B(T)$.
This is the \basequ setting in \tabref{compareresults} where
\basequ signifies that the LID is not given prior knowledge
about which languages occur in $T$.
For example, for the comparison of CLD3 and \modelname on \flores
in the top part (\basequ) of \tabref{compareresults}, both CLD3 and \modelname are run on
the entire \flores test set. We do not exclude
the languages that are present in $T$, but are not part of $M\dnrm{CLD3} \cap M\dnrm{GlotLID}$,
i.e., the languages
outside of the set of 95 languages common to CLD3 and
\modelname.

\textbf{Macro average.}
For a fair comparison to prior work,
we restrict 
the macro average over languages to a subset of languages in
order to replicate the experimental setup of this prior
work.
This subset is indicated in the tables.

\textbf{Realistic evaluation for low-resource scenarios.}
We believe that our new evaluation setup \basequ better
approximates real world situations. In cleaning pipelines, LID
models are often presented with an unknown set of languages
without prior knowledge. Therefore, it is crucial for an LID
to have the capacity to handle unknown languages. This
can be achieved by setting a threshold $\theta$ on the
confidence scores. If the confidence score for a predicted
label falls below the threshold, the model should label
the input text as ``undetermined''. This reduces the
risk of languages unknown to the model being incorrectly
categorized as a known language (the out-of-model problem).
Consequently, when comparing LIDs, it is necessary to apply
each model to the entire benchmark.

\subsection{Evaluation measures}\seclabel{measures}
Unlike some older prior work~\citep{jauhiainen2019automatic}, we do
not use accuracy because classes are highly imbalanced.
Instead, we follow recent prior work \citep{nllbteam2022language,burchell-etal-2023-open} and use F1 and false
positive rate (FPR). F1 is an aggregate measure of precision
and recall, both of which are important: we want accurate
classifications decisions (precision) and we do not want to
lose too much data (recall).
FPR is defined as \(\text{FPR} = \frac{{\text{FP}}}{{\text{FP} + \text{TN}}}\),
where \(\text{FP}\) is the number of false positives, and \(\text{TN}\) is the number of true negatives.
FPR helps us assess the
potentially fatal effect
of an even low false positive rate when the negative class is
huge -- which is the case in our scenario. For example, an
FPR of .01 (which \textsl{prima facie} may seem ok) for a
language $l$ with base frequency .01
can result in
a corpus for $l$ that contains 50\% noise, an unacceptably high
level.

\begin{table}[t]
\centering
\resizebox{1.0\linewidth}{!}{
\begin{tabular}{llr|ccccc}
&& & \multicolumn{2}{c}{\modelname, $\theta$=.0} & \multicolumn{2}{c}{\modelname, $\theta$=.5} \\
\cmidrule(lr){4-5} 
\cmidrule(lr){6-7}
\textbf{Benchmark}& & \boldsymbol{$|L|$} & \textbf{F1}$\uparrow$ & \textbf{FPR}$\downarrow$ & \textbf{F1}$\uparrow$ & \textbf{FPR}$\downarrow$\\
\midrule
\corpusname &all & 1832 & .940 & .0005 & .938 & .0003 \\
\corpusname &subset & 1665 & .977 & .0003 & .973 & .0002 \\
UDHR &all& 374 & .750 & .0015 & .734 & .0007 \\
UDHR & subset&342 & .784 & .0014 & .770 & .0006 \\
FLORES-200 &all& 196  & .917 & .0042 & .887 & .0013 \\
FLORES-200&subset & 177 & .957 & .0029 & .924 & .0010 \\
\end{tabular}
}
\caption{Performance of \modelname on  \corpusname, \udhr
  and \flores-200  test sets.
Subset: restriction to an ``operational'' subset of
languages that are either high-resource or
  for which \modelname achieves
F1$\neq$0 and FPR$\leq$.0005 on \corpusname test.
  $L$: 
intersection of \modelname languages
(all: 1832 or subset: 1665)
and languages present in benchmark.
Referring to \protect\figref{decisionrule}, the size of the base set $B$
is
either 1832 (all) or 1665 (subset).
$L$ is the
set of languages over which the macro average is computed.
For example, 
for the last line
(FLORES-200 subset),
$B$ consists of 1665
languages and the reported macro averages are computed over
177 languages.
\tablabel{glotlidresults}
}
\end{table}

\begin{table*}[t]
\centering
\resizebox{1.0\linewidth}{!}{
\begin{tabular}{lll|rrrrrrrr|rrrrrrrr}
&&& \multicolumn{8}{c}{\textbf{FLORES-200}} & \multicolumn{8}{|c}{\textbf{UDHR}} \\
&& 
& \multicolumn{2}{c}{\textbf{CLD3}}
& \multicolumn{2}{c}{\textbf{\ft176}}
& \multicolumn{2}{c}{\textbf{\edin}}
& \multicolumn{2}{c}{\textbf{\nllb}}
& \multicolumn{2}{|c}{\textbf{CLD3}}
& \multicolumn{2}{c}{\textbf{\ft176}}
& \multicolumn{2}{c}{\textbf{\edin}}
& \multicolumn{2}{c}{\textbf{\nllb}} \\
&& 
& \multicolumn{2}{c}{{$|L|$} = 96}
& \multicolumn{2}{c}{{$|L|$} = 108}
& \multicolumn{2}{c}{{$|L|$} = 195}
& \multicolumn{2}{c}{{$|L|$} = 188}
& \multicolumn{2}{|c}{{$|L|$} = 100}
& \multicolumn{2}{c}{{$|L|$} = 124}
& \multicolumn{2}{c}{{$|L|$} = 159}
& \multicolumn{2}{c}{{$|L|$} = 172}\\
&\textbf{LID Model} & \boldsymbol{$\theta$} & \textbf{F1}$\uparrow$ & \textbf{FPR}$\downarrow$ & \textbf{F1}$\uparrow$ & \textbf{FPR}$\downarrow$ & \textbf{F1}$\uparrow$ & \textbf{FPR}$\downarrow$ & \textbf{F1}$\uparrow$ & \textbf{FPR}$\downarrow$ & \textbf{F1}$\uparrow$ & \textbf{FPR}$\downarrow$ & \textbf{F1}$\uparrow$ & \textbf{FPR}$\downarrow$ & \textbf{F1}$\uparrow$ & \textbf{FPR}$\downarrow$ & \textbf{F1}$\uparrow$ & \textbf{FPR}$\downarrow$ \\\hline
\midrule
\multirow{6}{*}{\rotatebox{90}{\basequ}}
&baselines&.0         &.753&.0098&.775&.0090&.923&.0051& .947 & .0053
&.544&.0099&.566&.0079&.645&.0056& .641 & .0051 \\
&baselines&$\theta_1$ &.779&.0081& \underline{.816} &.0033&.923&.0050& .948  & .0051
&.576&.0081&.644&.0025&.676&.0046& .677 & .0040 \\
&baselines&$\theta_2$ &\underline{.799} &\underline{.0060}&.796&\textbf{.0021}& \textbf{.923} &\underline{.0044}& \textbf{.947}  & \underline{.0047}
&\underline{.618}&.\underline{0060}&\underline{.647}&\textbf{.0014}&\underline{.718}&\underline{.0034}& \underline{.717} & \underline{.0030} \\
&\modelname &.0   & .978 & .0051 & .987 & .0042 & \underline{.916} & .0043 & \textbf{.947} & .0035 & .868 & .0033 
 & .868 & .0030 & \textbf{.848} & .0020 & \textbf{.847} & .0019 \\
&\modelname &.3   & .980 & .0042 & .987 & .0037 &  .898 & .0020 & .927 & .0019  & .881 & .0028 
 & .879 & .0026 & .846 & .0015 & .844 & .0015 \\
&\modelname &.5   & \textbf{.980} & \textbf{.0031} & \textbf{.987} & \underline{.0029} & .886 & \textbf{.0014} & .916 & \textbf{.0013}  & \textbf{.903} & \textbf{.0023}
 & \textbf{.890} & \underline{.0021} & .847 & \textbf{.0012} & .846 & \textbf{.0011}\\\hline
\multirow{2}{*}{\rotatebox{90}{\baseex}}&baselines &.0  &\underline{.952} &\textbf{.0104} &\underline{.881} &\textbf{.0093} &\textbf{.923} &\textbf{.0051} & \underline{.950} & \textbf{.0053} &\underline{.922} &\underline{.0101} &\underline{.739} &\textbf{.0081} &\underline{.881} &\textbf{.0063} &\underline{.854} & \textbf{.0058}\\
&\modelname & .0 & \textbf{.983} & \textbf{.0104} & \textbf{.991} & \textbf{.0093} & \underline{.922} & \textbf{.0051} & \textbf{.954} & \textbf{.0053} & \textbf{.952} & \textbf{.0100} & \textbf{.927} & \textbf{.0081} & \textbf{.926} & \underline{.0064} & \textbf{.925} & \underline{.0060}

\end{tabular}
}
\caption{Evaluation of LID performance.
Top (``\basequ''): The set of languages is not known,
i.e., each LID makes predictions for all languages it was
trained on.
Bottom (``\baseex''): The set of languages is known:
each LID only makes predictions for languages that occur in
the benchmark.
For the more realistic ``\basequ'' setting, \modelname outperforms the baselines on \udhr (which
we take to be the best benchmark for the low-resource case)
assuming a good tradeoff between FPR and F1 is desired;
it either matches or outperforms them on \flores.
Let $M_i$ be the set of languages model $m_i$ was trained on
and $B(T)$ the set of languages covered by benchmark $T$.
Then
F1 and FPR are averages
over $L = M_1 \cap M_2 \cap B(T)$
when comparing models $m_1$
and $m_2$; this is indicated in the third row of table, e.g., 
$|L|=96$ for $m_1=$ CLD3, $m_2=$ GlotLID.
$\theta_1$=.5 for CLD3, $\theta_1$=.3 for \ft176, \edin and \nllb.
$\theta_2$=.7 for CLD3, $\theta_2$=.5 for \ft176, \edin and \nllb.
Referring to \protect\figref{decisionrule}, the base set $B$
in \basequ has size 
103  for CLD3, 176 for \ft176,
195 for \edin, 211 for
\nllb  and
 \numberlanguagestotal for \modelname (i.e., the languages
 the LID was trained on).
For scenario \baseex, $B=L$, i.e., $B=
 M_1 \cap M_2 \cap B(T)$. For example, 
$|B| = 96$ (for both CLD3 and GlotLID) for
the four cells in the
the \baseex
rows and the CLD3 columns in the lower left corner of the table. The best result in each column is \textbf{bolded}, and the second-best result is \underline{underlined}.
\tablabel{compareresults}
}
\end{table*}

\section{Results}

\seclabel{results}

\def\hcolsep{\hspace{0.095cm}}

\begin{table*}[t]
\centering
\tiny
\begin{tabular}{
l@{\hcolsep}||@{\hcolsep}l@{\hcolsep}r@{\hcolsep}r@{\hcolsep}l@{\hcolsep}r@{\hcolsep}r@{\hcolsep}
|l@{\hcolsep}r@{\hcolsep}r@{\hcolsep}l@{\hcolsep}r@{\hcolsep}r@{\hcolsep}
|l@{\hcolsep}r@{\hcolsep}r@{\hcolsep}l@{\hcolsep}r@{\hcolsep}r@{\hcolsep}
}
&\multicolumn{6}{c|}{\textbf{\flores-200}}
&
\multicolumn{6}{c|}{\textbf{\udhr}}
&
\multicolumn{6}{c}{\textbf{{\corpusname}}}
\\
&\textbf{language}& \textbf{FP} & \textbf{cl}& \textbf{top FP source} & \textbf{\#FP} & \textbf{\%}
&
\textbf{language}& \textbf{FP} & \textbf{cl}& \textbf{top FP source} & \textbf{\#FP} & \textbf{\%}
&
\textbf{language} & \textbf{FP} & \textbf{cl}& \textbf{top FP source} & \textbf{\#FP} & \textbf{\%}
\\\hline\hline
\multirow{5}{*}{\rotatebox{90}{{most errors}}}
\tablebreak
&	arb:St Arabic & 3787 & .18 & ars:Najdi Arabi & 829 & .22	&	cmn:Mandarin Ch & 596 & .38 & chr:Cherokee & 81 & .14	&	spa:Spanish & 1952 & .34 & pid:Piaroa & 156 & .08	\\
&	arz:Egyptian Ar & 1726 & .32 & apc:Levantine A & 440 & .25	&	qub:Huallaga Hu & 247 & .00 & qvh:Huamalíes-D & 55 & .22	&	eng:English & 1168 & .46 & lir:Liberian En & 254 & .22	\\
&	pes:Ir. Persian & 1495 & .40 & prs:Dari & 905 & .61	&	fin:Finnish & 224 & .22 & krl:Karelian & 138 & .62	&	rus:Russian & 1057 & .49 & chu:Church Slav & 661 & .63	\\
&	cmn:Mandarin Ch & 1008 & .00 & yue:Yue Chinese & 1008 & .99	&	wuu:Wu Chinese & 172 & .24 & hak:Hakka Chine & 44 & .26	&	bho:Bhojpuri & 882 & .50 & bih:Bihari Lgs & 854 & .97	\\
&	hin:Hindi & 977 & .51 & awa:Awadhi & 693 & .71	&	rus:Russian & 157 & .28 & niv:Gilyak & 44 & .28	&	lir:Liberian En & 712 & .47 & din:Dinka & 174 & .24	\\\hline

\multirow{5}{*}{\rotatebox{90}{{most noisy}}}
\tablebreak
&	arb:St Arabic & 3787 & .18 & ars:Najdi Arabi & 829 & .22	&	evn:Evenki & 36 & .23 & oaa:Orok & 19 & .53	&	rus:Russian & 1057 & .49 & chu:Church Slav & 661 & .63	\\
&	arz:Egyptian Ar & 1726 & .32 & apc:Levantine A & 440 & .25	&	quz:Cusco Quech & 82 & .40 & qxu:Arequipa-La & 61 & .74	&	eng:English & 1168 & .46 & lir:Liberian En & 254 & .22	\\
&	prs:Dari & 338 & .24 & pbt:S Pashto & 310 & .92	&	hrv:Croatian & 84 & .42 & bos:Bosnian & 39 & .46	&	spa:Spanish & 1952 & .34 & pid:Piaroa & 156 & .08	\\
&	dyu:Dyula & 255 & .25 & bam:Bambara & 255 & .99	&	tzm:C Atlas Tam & 52 & .02 & zgh:St Moroccan & 52 & .99	&	crq:Iyo'wujwa C & 347 & .47 & crt:Iyojwa'ja C & 347 & .99	\\
&	apc:Levantine A & 161 & .42 & ajp:S Levantine & 70 & .43	&	uzn:N Uzbek & 72 & .46 & cbu:Candoshi-Sh & 16 & .22	&	crt:Iyojwa'ja C & 698 & .48 & crq:Iyo'wujwa C & 697 & .99 
\\\hline

\multirow{5}{*}{\rotatebox{90}{{no positives}}}
\tablebreak
&	&&&&&	&	tet:Tetum & 0 & .00 &  &  & 	&	sck:Sadri & 0 & .00 &  &  & 	\\
&	&&&&&	&	hsn:Xiang Chine & 0 & .00 &  &  & 	&	chg:Chagatai & 0 & .00 &  &  & 	\\
&	&&&&&	&	abk:Abkhazian & 0 & .00 &  &  & 	&	liv:Liv & 0 & .00 &  &  & 	\\
&	&&&&&	&	vep:Veps & 0 & .00 &  &  & 	&	gbm:Garhwali & 0 & .00 &  &  & 	\\
&	&&&&&	&	niv:Gilyak & 0 & .00 &  &  & 	&	tmw:Temuan & 0 & .00 &  &  & 	\\\hline

\multirow{5}{*}{\rotatebox{90}{{hi resource}}}
\tablebreak
&	arb:St Arabic & 3787 & .99 & ars:Najdi Arabi & 829 & .22	&	cmn:Mandarin Ch & 596 & .99 & chr:Cherokee & 81 & .14	&	cmn:Mandarin Ch & 367 & .99 & wuu:Wu Chinese & 208 & .57	\\
&	dzo:Dzongkha & 10300 & .09 & bod:Tibetan & 10300 & .99	&	fin:Finnish & 224 & .99 & krl:Karelian & 138 & .62	&	eng:English & 1267 & .99 & lir:Liberian En & 254 & .20	\\
&	hin:Hindi & 977 & .99 & awa:Awadhi & 693 & .71	&	hin:Hindi & 76 & .99 & mai:Maithili & 24 & .32	&	hin:Hindi & 488 & .99 & bho:Bhojpuri & 98 & .20	\\
&	rus:Russian & 1 & .99 & bul:Bulgarian & 1 & .99	&	rus:Russian & 256 & .99 & eng:English & 100 & .39	&	rus:Russian & 1156 & .99 & chu:Church Slav & 661 & .57	\\
&	spa:Spanish & 10 & .99 & ast:Asturian & 7 & .70	&	spa:Spanish & 62 & .99 & agr:Aguaruna & 20 & .32	&	spa:Spanish & 1952 & .99 & pid:Piaroa & 156 & .08	
\end{tabular}
\caption{Analysis of the \modelname runs with settings
$\theta$=.0, \basequ from \protect\tabref{glotlidresults}
and \protect\tabref{compareresults}. ``most errors'':
languages with the most false positives. ``most noisy'': a
sample of languages with cleanness between 0 and .5. ``no
positives'': a sample of languages without positives. ``hi
resource'': a more realistic setting in which the
distribution is skewed in favor of high-resource languages.
For each ``language'', we give the number of false positives
(``FP''), the cleanness of the resulting corpus (``cl'':
ratio true positives to all positives), its most conflated
language (``top FP source''), FP contributed by that
language and 
the ratio of the two FP numbers (``\%'').
To
save space, we write .99 for 1.00.
\tablabel{floresfalsepositive}}
\end{table*}

\tabref{glotlidresults} gives results on \corpusname test,
UDHR and FLORES-200.
\modelname does not perform well on some
languages. In particular, there are \numberlanguagediff
(\numberlanguagestotal-\numberlanguagesaccurate) low-resource languages
for which either F1$<$.01 or FPR$>$.0005, often due
to very small \corpusname training sets.
The table gives results for ``all'' \numberlanguagestotal languages as well as for
the ``subset'' of \numberlanguagesaccurate well-performing languages.
We run \modelname in two settings: $\theta$=.0 (i.e.,
we choose the highest probability class no matter how low
its probability is) and $\theta =.5$ (i.e., we only assign a
language label if its probability
exceeds~$.5$). See \figref{decisionrule} for the
definition of our decision rule.

Focusing on the ``subset'' results for $\theta =.5$,
F1 is .973 on \corpusname and .924 on \flores; and FPR is
.0002 on \corpusname and .0010 on \flores. This is a very good
performance, in particular for the use case of low-resource
corpus creation because low FPR means that the
resulting corpora will be less contaminated. 
 On \udhr, again for the ``subset'' results for $\theta =.5$, F1 is
.770 and FPR .0006. This is again an encouragingly low FPR,
but F1 is quite a bit lower than for \corpusname
and \flores. The reason is that we have a domain shift
(compared to \corpusname) and many more languages (compared
to \flores), resulting in lower F1. Although the \udhr
results should be improved further, we will now show that
they outperform the state of the art.

\tabref{compareresults} compares \modelname with four baselines.
We consider two evaluation settings
(\basequ and \baseex)
and three thresholds $\theta$.
The top part of the table (\basequ) corresponds to the
case where the set  of languages in the benchmark is not known,
i.e., the LID makes predictions for all languages it was
trained on.
In contrast, in the
\baseex setting (bottom part), the set of languages in the
benchmark is known, and
each LID only makes predictions for those languages.
\basequ is a more realistic setting, as we usually do not know which
languages occur in a corpus that needs to be cleaned.

For the \basequ setting, \modelname consistently outperforms
CLD3 by a large margin. Taking into account that F1 and FPR
should be balanced, we also take it to outperform \ft176.
Even though \modelname's FPR is slightly higher in some
cases, its F1 is better by a large margin, so
that it is clearly the better performing system.

On \udhr, \modelname also clearly outperforms \edin
and \nllb for F1 and FPR by large margins. On \flores, F1 is
slightly worse and FPR slightly better compared with \edin
and \nllb.
We point out that this comparison is not entirely
fair since \edin and \nllb were designed with \flores in
mind. More
importantly, our use case is the creation of low-resource
corpora for which \udhr is the more appropriate benchmark.

Comparing results for different
thresholds, we observe
that increasing $\theta$  lowers F1 (because
recall is hurt) and lowers FPR (because precision is increased).
This suggests that a higher threshold should be used since
lower FPR will result in low-resource corpora with less
contamination from high-resource languages.

For the less realistic \baseex setting,
\modelname performs better than CLD3 and \ft176 and
comparably to \edin and \nllb. 
Overall, \modelname  clearly outperforms all baselines
for the low-resource corpus creation use case.

To analyze variance of results, we ran
three GlotLID experiments with different initial seeds
on the 200 languages with the most data, splitting the data
into 80\% train and 20\% test. The F1 score was .991 each
time. This indicates that the variance of FastText in this task (and by
extension GlotLID) is negligible.

\section{Analysis}
\seclabel{analysis}

In this section, we analyze
the \modelname
results summarized
in \tabref{glotlidresults}
($\theta$=.0, ``all'')
for our
main use case, the creation of high-quality
corpora. We address four questions. (i) For which languages
do we get a high  number of false positives? (ii)
For which languages do we produce a corpus with a high
contamination rate?
(iii) For which
languages does learning completely fail? (iv)
Is it more realistic to evaluate LID on a balanced test set
(as in prior work) or on one that is skewed in favor
of high-resource languages?

\textbf{Most errors.}
We first analyze
languages with a high number of errors.
\tabref{floresfalsepositive} (top, ``most errors'')
gives
for each of the three
benchmarks 
the five languages that have the highest number of errors
(column ``language''). ``FP'' is the number of false
positives, ``cl'' the ratio of true positives to all
positives (that is the ``cleanness'' of the corpus), ``top
FP source'' the language that contributed most of the errors
and ``\%'' is the portion of these false positives as a
percentage of all false positives.
We use the cl measure in our analysis
because it is ultimately the measure we want to optimize to
produce high-quality low-resource corpora.
Note that cl (the denominator is the total  number of positive sentences) is
not directly related to FPR (the denominator is the number of
sentences  that do not belong to the language). cl is a more
direct measure of the utility of the resulting corpus of a
low-resource language (e.g., for training a language
model) than FPR.

Most of the fifteen pairs of ``conflated'' languages
shown in the table are
closely related languages: varieties of Arabic (Standard,
Najdi, Egyptian and Levantine), Persian (Iranian, Dari), Chinese (Mandarin, Yue,
Wu, Hakka), English (Standard, Liberian), Quechua (Huallaga
Huánuco, Huamalíes-Dos de Mayo Huánuco), Finnic (Finnish,
Karelian), Slavic (Russian, Church Slavic), Bihari
(Bihari, Bhojpuri) and Hindi (Standard, Awadhi).  In many of
these cases, speakers of one variety of the pair also have
good knowledge of the other; e.g., many speakers of Arabic
varieties know Standard Arabic. The two Quechua varieties
are spoken in neighboring areas of Peru. The quantitatively
largest use of Church Slavic  (which may be reflected in the
size of our corpora) is in Russia by Russian speakers.

Arabic, Chinese and English (and perhaps also Hindi,
Persian and Bihari) are diglossic linguistic communities.
There may be a lack of  clear separation between
the two conflated varieties  in the available corpora because
speakers switch back and forth between more formal and less
formal ways of speaking depending on factors like context,
audience and subject.
This type of fluid switching between languages often occurs
in a single sentence or conversation, i.e., it manifests as code switching.
As a result, much of the text (and
speech) produced in one language may be mixed with the
other language.
New methods will have to be developed
to deal with these quite complex challenges of
creating training corpora for
language identification; see also \citep{aguilar-etal-2020-lince}.

Apart from these related languages, at least four
conflated language pairs
in \tabref{floresfalsepositive}
are clear errors:
Mandarin/Cherokee, Russian/Gilyak, Spanish/Piaroa and
Liberian English/Dinka.
Similar to the situation we described for the closely
related languages,
Gilyak (resp.\ Piaroa) is spoken in an area where Russian
(resp.\ Spanish)
is the dominant
official language.
This means that our training corpora will need to be
improved: they most likely contain many
sentences labeled as Gilyak/Piaroa that are partially or
completely Russian/Spanish. 
We leave it to future work to
revisit and improve our corpus selection and preprocessing
methodology to address this data quality problem.

\modelname confuses
Mandarin and Cherokee because
our Cherokee training data do not cover
the Cherokee syllabary script.
Sentences written in this script are returned with a
close to uniform distribution over several other scripts,
including Chinese, Japanese and Thai, which explains the confusion.
The Dinka test set is noisy. In a manual inspection, we
found 377 sentences that are clearly English, not
Dinka. Because \modelname did not learn very well to discriminate
English and Liberian English, 174 of these 377 sentence were
classified as Liberian English.

\textbf{Most noisy corpora.}
The second part of 
\tabref{floresfalsepositive} (``most noisy'') gives,
for each benchmark, a random selection of five languages
whose cleanness score cl (ratio of true positive to all
positives) is in the range 0$<$cl$<$.5. The total number of
languages in this range is 9 for \flores, 27 for \udhr and 6
for \corpusname. Again, most of the conflated pairs are
closely related languages as in the last section. Additional
pairs that occur here are Dyula/Bambara, Evenki/Orok,
Croation/Bosnian, Berber languages (Standard Moroccan
Tamazight,
Atlas
Tamazight) and two varieties of Chorote (Iyo'wujwa,
Iyojwa’ja). The resulting corpora are noisy,
an issue that we will have to address in future work.

\textbf{No positives.}
Part 3 of \tabref{floresfalsepositive} (``no positives'') gives five random examples from
languages for which there was not a single positive
classification. There were no such languages for \flores.

For \udhr, we identified two reasons. (i) Performance
on \corpusname is good, but poor on \udhr. Tetum is an
example.  The most likely cause is a domain shift or some
other big train/test difference.  (ii) The training set
is too small (less than 30 sentences): hsn (Xiang Chinese),
abk (Abkhazian), vep (Veps) and niv (Gilyak) are in this
class.

For the five \corpusname random examples with no positives,
the reason is also that the training sets
were too small (less than 40 sentences): sck (Sadri), chg
(Chagatai), liv (Liv), gbm (Garhwali) and tmw (Temuan). We should have set a higher threshold for
minimum size of the training corpus. Note that the number
of \numberlanguagesaccurate languages that we use throughout
the paper already reflects this insight. Even though we
train on
\numberlanguagestotal languages, we claim
reasonable performance for only
\numberlanguagesaccurate (\tabref{glotlidresults}).

\textbf{Test set skewed in favor
of high-resource.}
\flores and \udhr test sets are balanced: high-resource and
low-resource languages have about the same size. Following
this model, we constructed the test set of \corpusname
in the same way. F1 is independent of this distribution, but
FPR and cleanness (``cl'') are strongly dependent on it. The
Spanish corpus generated by \modelname on \corpusname test
has a dismal cleanness of only .34. Is this a problem
for \modelname?

We believe the answer is no, as the
corpora we run LID on will have a distribution skewed
in favor of high-resource languages.
To simulate this more realistic scenario,
the last part of \tabref{floresfalsepositive}
(``hi resource'')
gives five selected languages for each benchmark
where we have inflated the subsets for high-resource
languages by a factor of 100. For example, instead of a
single copy of the English part of \flores, the test set now
contains 100 copies.

We see in \tabref{floresfalsepositive} that this results in
clean corpora (cl=.99) for each of
the fourteen
high-resource languages shown: Standard Arabic, Hindi,
Russian, Spanish (\flores); Mandarin, Finnish, Hindi,
Russian, Spanish (\udhr); Mandarin, English, Hindi, Russian,
Spanish (\corpusname). As an example, looking at Spanish for
\corpusname (the first
and last lines in the table), the number of false positives
(1952) and the number of false positives contributed by the
low-resource language Piaroa (156) are the same. But since
the size of Spanish is increased 100x, its
cleanness improves from .34 for the unrealistic
uniform distribution to .99 for the realistic skewed
distribution. Thus, as we would
expect, LID for high-resource languages is a relatively  easy
problem and this does not change much if we run a
broad-coverage LID like \modelname.

Conversely, LID numbers for low-resource languages can be
\emph{negatively} affected.
The Dzongkha corpus generated from \flores
in the uniform setting has 103 false positives and a
cleanness of .91 (not shown). In the skewed setting, 
making Tibetan a high-resource language causes 10,300 false
positives from Tibetan to leak into Dzongkha, reducing its cleanness to an unacceptable .09. 

This discussion suggests that the established evaluation
methodology for LID is unsatisfactory. We recommend that
future work considers both unifom and skewed test sets to
better assess how LID is expected to perform in the real world.

This analysis  demonstrates
how much harder LID becomes
when we represent as large and diverse sets of languages as we do. What
we have shown is that there is a real danger of creating
corpora that are badly contaminated.
To address this, we
need to develop methodologies and resources that better handle
low-resource languages.

Based on the analysis described in this section we created
and open-sourced a much improved version of the UDHR test
set for evaluation of
LID.\footnote{\url{https://huggingface.co/datasets/cis-lmu/udhr-lid}}
All UDHR results in this paper are based on the version of
the UDHR test set descibed in \secref{evaluation_data}.

\section{Conclusion}
We create \corpusname, an  LID resource that
covers \numberlanguagestotal languages, several times more than prior
work.
We introduce \modelname, an open-source LID that
covers \numberlanguagesaccurate languages with good results.
The comparison of \modelname
against four  LID baselines shows
superior performance
for the low-resource use case. 
In future research, we would like to improve quality of our training corpora and add more low-resource languages in to \genericname.
We hope \genericname will
be a valuable
resource in creating higher-quality corpora for low-resource languages.

\section*{Limitations}

(1) We publish list of \corpusname data sources 
as part of this work. There is no other LID benchmark available that
covers as many languages as \corpusname does. \corpusname,
\flores and \udhr all have drawbacks as evaluation datasets
for LID. An LID trained on \corpusname train and tested
on \corpusname test will often find the same domain in the
test set as in the training set. It is well known that this
results in overly optimistic evaluation numbers.
\flores and \udhr consist of data that were not originallly
produced in each language.
Rather, they were translated from
high-resource languages.
The same is true to a lesser extent for \corpusname.
Translated language is only an
imperfect evaluation benchmark because it can differ greatly
from natural language data, i.e., translationese is
often not a good model of  natural language data.

(2) Many corpora for the lowest resource languages are
derived from religious sources. It should be noted that many
Bible translations do not reflect actual language use.

(3) We do not conduct hyperparameter search and instead use the hyperparameters employed by previous studies. However, conducting such a search can make our findings more robust, considering the difference in the number of languages included in our study compared to the prior work.

(4) Although we tried our best to select the most suitable LIDs as the baseline. We could not compare against all of the LID models. This includes CLD2~\citep{mccandless2010accuracy},
Equilid~\citep{jurgens-etal-2017-incorporating}, 
Langdetect~\citep{shuyo2010language},
langid.py~\citep{lui-baldwin-2012-langid}, whatlang \citep{brown-2014-non, la-strings}, idNet
\citep{dunn2020mapping}, Franc~\citep{franc}, AfroLID~\citep{adebara-etal-2022-afrolid}, 
HeLI-OTS~\citep{jauhiainen-etal-2022-heli},
transliterate~\citep{transliterate}, whatlang-rs
~\citep{whatlang-rs},
lingua~\citep{lingua}, Google/Bing Online, LanideNN~\citep{kocmi-bojar-2017-lanidenn}, Paasaa~\citep{paasaa}, Q-LID~\citep{ren-etal-2022-effective}, UDLDI~\citep{goswami-etal-2020-unsupervised}, PALI~\citep{ahmadi-etal-2023-pali}, SS-LID~\citep{caswell-etal-2020-language, bapna2022building, kudugunta2023madlad}
and
TextCat~\citep{cavnar1994n}.

\section*{Ethics Statement}
We here highlight key ethical considerations for
\genericname.

\textbf{Data.}
The data used in our study comes from openly available
(but not necessarily freely redistributable)
datasets, including resources previously published by
researchers, publishers, and translators. We ensured that
the data collection process complied with licensing of each
dataset.

\textbf{Bias.} 
We recognize potential biases towards higher resource languages. We conducted a comprehensive analysis of errors and evaluated their impact on our results.

\textbf{Inclusivity.}
We acknowledge the challenges associated with low-resource languages and have taken steps to include a diverse range of languages in our study.

\textbf{Ethical Use.}
We have demonstrated both positive and negative outcomes of
applying \modelname as an LID tool. We acknowledge
that \modelname
has a high error rate for some low-resource languages.
This means that there is a  potential risk of
excluding low-resource languages during the collection and
processing of NLP corpora.

\textbf{Transparency.}
We provide detailed descriptions of our methodology, model
architecture, and evaluation process. Additionally, we  make
our research artifacts, including model, code, and list of data sources openly available
to foster collaboration and reproducibility.

\section{Acknowledgements}

We would like to thank anonymous reviewers.
This work was funded by the European Research
Council (grant \#740516).

\bibliography{anthology,custom}

\appendix

\section{Evaluation data issues}

\subsection{\flores-200}\seclabel{flores-issues}

There are some mistakes in the \flores-200 dataset which have been raised by the community. 

For example, in a GitHub issue \url{https://github.com/facebookresearch/flores/issues/61}, it is pointed out that yue\_Hant and zho\_Hant should actually be very easy to distinguish from each other, and the Cantonese (Yue Chinese, yue\_Hant) data in \flores-200 is completely wrong.

In another issue (\url{https://github.com/facebookresearch/flores/issues/63}), it is mentioned that the Central Atlas Tamazight (tzm) is actually in Standard Moroccan Tamazight (zgh), as confirmed by a native speaker of Central Atlas Tamazight.

\subsection{UDHR}\seclabel{udhr-issues}

There are some mistakes with UDHR. For example, both ckb and kmr files are the same. ckb is known for the Arabic script, although it can also be written in Latin. There are also some files that the writing system is not in popular use (based on \citet{kargaran2023glotscript} metadata):

\begin{itemize}[noitemsep]
  \item ckb\_Latn (Arabic script is in use.)
  \item azb\_Latn (Arabic script is in use.)
  \item khk\_Mong (Cyrillic script is in use.)
  \item vie\_Hani (Latin script is in use.)
\end{itemize}

\section{Performance of \modelname per language}\seclabel{appendix_glot_performance}

The list of languages used to train \modelname, along with the corresponding amount of available data and detailed results for each language, can be found in
Tables~\ref{tab:appendix_glot_per_lang_1}-\ref{tab:appendix_glot_per_lang_25}

\begin{table*}[ht]
\small
\centering
\resizebox{0.57\textheight}{!}{
\begin{tabular}{llrrrrrrr}
& 
& 
& \multicolumn{2}{c}{\textbf{\corpusname}}
& \multicolumn{2}{c}{\textbf{\flores-200}} 
& \multicolumn{2}{c}{\textbf{\udhr}} 
\\
\cmidrule(lr){4-5} 
\cmidrule(lr){6-7}
\cmidrule(lr){8-9}
\textbf{iso639-3} &
\textbf{Language} &
\textbf{$|$Sentences$|$} &
\textbf{F1}$\uparrow$ & \textbf{FPR}$\downarrow$ &
\textbf{F1}$\uparrow$ &
\textbf{FPR}$\downarrow$ &
\textbf{F1}$\uparrow$ & 
\textbf{FPR}$\downarrow$ 
\\
\midrule
aai         & Arifama-Miniafia         & 7954         & 0.9995         & 0.0         &          &          &          &          \\

aak         & Ankave         & 7784         & 1.0         & 0.0         &          &          &          &          \\

aau         & Abau         & 7888         & 1.0         & 0.0         &          &          &          &          \\

aaz         & Amarasi         & 9071         & 0.999         & 0.0         &          &          &          &          \\

aba         & Abé         & 28725         & 0.81648         & 0.00371         &          &          &          & 0.00416         \\

abk         & Abkhazian         & 28         &          &          &          &          &          &          \\
\end{tabular}
    }
    \caption{Performance of \modelname on \corpusname test, \flores-200 and \udhr benchmarks (part 1)}\label{tab:appendix_glot_per_lang_1}\end{table*}
\begin{table*}[h]
\small
\centering
\resizebox{0.57\textheight}{!}{
\begin{tabular}{llrrrrrrr}
& 
& 
& \multicolumn{2}{c}{\textbf{\corpusname}}
& \multicolumn{2}{c}{\textbf{\flores-200}} 
& \multicolumn{2}{c}{\textbf{\udhr}} 
\\
\cmidrule(lr){4-5} 
\cmidrule(lr){6-7}
\cmidrule(lr){8-9}
\textbf{iso639-3} &
\textbf{Language} &
\textbf{$|$Sentences$|$} &
\textbf{F1}$\uparrow$ & \textbf{FPR}$\downarrow$ &
\textbf{F1}$\uparrow$ &
\textbf{FPR}$\downarrow$ &
\textbf{F1}$\uparrow$ & 
\textbf{FPR}$\downarrow$ 
\\
\midrule
amk         & Ambai         & 7850         & 0.9995         & 3e-05         &          &          &          &          \\

amm         & Ama (Papua New Guinea)         & 7826         & 1.0         & 0.0         &          &          &          &          \\

amn         & Amanab         & 14975         & 0.9995         & 3e-05         &          &          &          & 0.00011         \\

amp         & Alamblak         & 7720         & 0.9985         & 3e-05         &          &          &          &          \\

amr         & Amarakaeri         & 7671         & 1.0         & 0.0         &          &          & 1.0         & 0.0         \\

amu         & Guerrero Amuzgo         & 8056         & 0.999         & 0.0         &          &          &          & 0.00022         \\
\end{tabular}
    }
    \caption{Performance of \modelname on \corpusname test, \flores-200 and \udhr benchmarks (part 2)}\label{tab:appendix_glot_per_lang_2}\end{table*}
\begin{table*}[ht]
\small
\centering
\resizebox{0.54\textheight}{!}{
\begin{tabular}{llrrrrrrr}
& 
& 
& \multicolumn{2}{c}{\textbf{\corpusname}}
& \multicolumn{2}{c}{\textbf{\flores-200}} 
& \multicolumn{2}{c}{\textbf{\udhr}} 
\\
\cmidrule(lr){4-5} 
\cmidrule(lr){6-7}
\cmidrule(lr){8-9}
\textbf{iso639-3} &
\textbf{Language} &
\textbf{$|$Sentences$|$} &
\textbf{F1}$\uparrow$ & \textbf{FPR}$\downarrow$ &
\textbf{F1}$\uparrow$ &
\textbf{FPR}$\downarrow$ &
\textbf{F1}$\uparrow$ & 
\textbf{FPR}$\downarrow$ 
\\
\midrule
bas         & Basa (Cameroon)         & 132504         & 0.996         & 0.00014         &          & 0.00017         &          & 0.00022         \\

bav         & Vengo         & 7954         & 1.0         & 0.0         &          &          &          & 0.00011         \\

bba         & Baatonum         & 33686         & 0.999         & 6e-05         &          &          & 0.92187         & 0.00099         \\

bbb         & Barai         & 10264         & 1.0         & 0.0         &          &          &          &          \\

bbc         & Batak Toba         & 279942         & 0.98618         & 0.00074         &          &          &          &          \\

bbj         & Ghomálá'         & 27499         & 0.79298         & 0.00465         &          &          &          & 0.00109         \\
\end{tabular}
    }
    \caption{Performance of \modelname on \corpusname test, \flores-200 and \udhr benchmarks (part 3)}\label{tab:appendix_glot_per_lang_3}\end{table*}
\begin{table*}[h]
\small
\centering
\resizebox{0.56\textheight}{!}{
\begin{tabular}{llrrrrrrr}
& 
& 
& \multicolumn{2}{c}{\textbf{\corpusname}}
& \multicolumn{2}{c}{\textbf{\flores-200}} 
& \multicolumn{2}{c}{\textbf{\udhr}} 
\\
\cmidrule(lr){4-5} 
\cmidrule(lr){6-7}
\cmidrule(lr){8-9}
\textbf{iso639-3} &
\textbf{Language} &
\textbf{$|$Sentences$|$} &
\textbf{F1}$\uparrow$ & \textbf{FPR}$\downarrow$ &
\textbf{F1}$\uparrow$ &
\textbf{FPR}$\downarrow$ &
\textbf{F1}$\uparrow$ & 
\textbf{FPR}$\downarrow$ 
\\
\midrule
bqj         & Bandial         & 7903         & 0.998         & 3e-05         &          &          &          &          \\

bqp         & Busa         & 7894         & 0.9955         & 0.00014         &          &          &          & 0.00011         \\

bre         & Breton         & 16810         & 0.98688         & 0.00011         &          &          & 0.98361         & 0.0         \\

bru         & Eastern Bru         & 7888         & 0.999         & 0.0         &          &          &          &          \\

brx         & Bodo (India)         & 11         &          &          &          &          &          &          \\

bsc         & Bassari         & 7949         & 0.9995         & 3e-05         &          &          &          & 0.00022         \\
\end{tabular}
    }
    \caption{Performance of \modelname on \corpusname test, \flores-200 and \udhr benchmarks (part 4)}\label{tab:appendix_glot_per_lang_4}\end{table*}
\begin{table*}[h]
\small
\centering
\resizebox{0.56\textheight}{!}{
\begin{tabular}{llrrrrrrr}
& 
& 
& \multicolumn{2}{c}{\textbf{\corpusname}}
& \multicolumn{2}{c}{\textbf{\flores-200}} 
& \multicolumn{2}{c}{\textbf{\udhr}} 
\\
\cmidrule(lr){4-5} 
\cmidrule(lr){6-7}
\cmidrule(lr){8-9}
\textbf{iso639-3} &
\textbf{Language} &
\textbf{$|$Sentences$|$} &
\textbf{F1}$\uparrow$ & \textbf{FPR}$\downarrow$ &
\textbf{F1}$\uparrow$ &
\textbf{FPR}$\downarrow$ &
\textbf{F1}$\uparrow$ & 
\textbf{FPR}$\downarrow$ 
\\
\midrule
chf         & Tabasco Chontal         & 8213         & 0.998         & 6e-05         &          &          &          & 0.00022         \\

chg         & Chagatai         & 8         &          &          &          &          &          &          \\

chj         & Ojitlán Chinantec         & 6824         & 0.72469         & 0.00091         &          &          & 0.12727         & 0.00471         \\

chk         & Chuukese         & 377198         & 0.99701         & 0.00017         &          &          & 0.97521         & 0.00011         \\

chn         & Chinook jargon         & 53         & 0.8         & 0.0         &          &          &          &          \\

cho         & Choctaw         & 124         & 0.78788         & 0.0         &          &          &          & 0.00263         \\
\end{tabular}
    }
    \caption{Performance of \modelname on \corpusname test, \flores-200 and \udhr benchmarks (part 5)}\label{tab:appendix_glot_per_lang_5}\end{table*}
\begin{table*}[h]
\small
\centering
\resizebox{0.57\textheight}{!}{
\begin{tabular}{llrrrrrrr}
& 
& 
& \multicolumn{2}{c}{\textbf{\corpusname}}
& \multicolumn{2}{c}{\textbf{\flores-200}} 
& \multicolumn{2}{c}{\textbf{\udhr}} 
\\
\cmidrule(lr){4-5} 
\cmidrule(lr){6-7}
\cmidrule(lr){8-9}
\textbf{iso639-3} &
\textbf{Language} &
\textbf{$|$Sentences$|$} &
\textbf{F1}$\uparrow$ & \textbf{FPR}$\downarrow$ &
\textbf{F1}$\uparrow$ &
\textbf{FPR}$\downarrow$ &
\textbf{F1}$\uparrow$ & 
\textbf{FPR}$\downarrow$ 
\\
\midrule
cwt         & Kuwaataay         & 7828         & 1.0         & 0.0         &          &          &          &          \\

cya         & Nopala Chatino         & 7946         & 0.82102         & 0.00891         &          &          &          & 0.00033         \\

cym         & Welsh         & 273301         & 0.99501         & 0.00019         & 0.99951         & 6e-05         & 1.0         & 0.0         \\

czt         & Zotung Chin         & 9478         & 0.999         & 3e-05         &          &          &          & 0.00011         \\

daa         & Dangaléat         & 7839         & 0.999         & 3e-05         &          &          &          & 0.00011         \\

dad         & Marik         & 7726         & 0.9995         & 3e-05         &          &          &          &          \\
\end{tabular}
    }
    \caption{Performance of \modelname on \corpusname test, \flores-200 and \udhr benchmarks (part 6)}\label{tab:appendix_glot_per_lang_6}\end{table*}
\begin{table*}[h]
\small
\centering
\resizebox{0.57\textheight}{!}{
\begin{tabular}{llrrrrrrr}
& 
& 
& \multicolumn{2}{c}{\textbf{\corpusname}}
& \multicolumn{2}{c}{\textbf{\flores-200}} 
& \multicolumn{2}{c}{\textbf{\udhr}} 
\\
\cmidrule(lr){4-5} 
\cmidrule(lr){6-7}
\cmidrule(lr){8-9}
\textbf{iso639-3} &
\textbf{Language} &
\textbf{$|$Sentences$|$} &
\textbf{F1}$\uparrow$ & \textbf{FPR}$\downarrow$ &
\textbf{F1}$\uparrow$ &
\textbf{FPR}$\downarrow$ &
\textbf{F1}$\uparrow$ & 
\textbf{FPR}$\downarrow$ 
\\
\midrule
epo         & Esperanto         & 1161088         & 0.98377         & 0.00091         & 0.99852         & 0.00017         & 0.96825         & 0.00044         \\

eri         & Ogea         & 7908         & 0.9995         & 3e-05         &          &          &          &          \\

ese         & Ese Ejja         & 7842         & 0.9995         & 0.0         &          &          & 0.75         & 0.00011         \\

esi         & North Alaskan Inupiatun         & 7915         & 0.992         & 0.00022         &          &          &          & 0.00011         \\

esk         & Northwest Alaska Inupiatun         & 7921         & 0.99249         & 0.00017         &          &          &          & 0.00011         \\

est         & Estonian         & 3664155         & 0.99399         & 0.00011         & 1.0         & 0.0         &          &          \\
\end{tabular}
    }
    \caption{Performance of \modelname on \corpusname test, \flores-200 and \udhr benchmarks (part 7)}\label{tab:appendix_glot_per_lang_7}\end{table*}
\begin{table*}[h]
\small
\centering
\resizebox{0.56\textheight}{!}{
\begin{tabular}{llrrrrrrr}
& 
& 
& \multicolumn{2}{c}{\textbf{\corpusname}}
& \multicolumn{2}{c}{\textbf{\flores-200}} 
& \multicolumn{2}{c}{\textbf{\udhr}} 
\\
\cmidrule(lr){4-5} 
\cmidrule(lr){6-7}
\cmidrule(lr){8-9}
\textbf{iso639-3} &
\textbf{Language} &
\textbf{$|$Sentences$|$} &
\textbf{F1}$\uparrow$ & \textbf{FPR}$\downarrow$ &
\textbf{F1}$\uparrow$ &
\textbf{FPR}$\downarrow$ &
\textbf{F1}$\uparrow$ & 
\textbf{FPR}$\downarrow$ 
\\
\midrule
gkp         & Guinea Kpelle         & 45287         & 0.54231         & 0.01537         &          &          & 0.92063         & 0.00109         \\

gla         & Scottish Gaelic         & 95320         & 0.99298         & 0.00011         & 0.99951         & 6e-05         & 0.94574         & 0.00022         \\

gle         & Irish         & 274216         & 0.9935         & 0.00017         & 1.0         & 0.0         & 0.95         & 0.00088         \\

glg         & Galician         & 343533         & 0.98495         & 0.00033         & 0.99703         & 0.00011         & 0.98305         & 0.00011         \\

glk         & Gilaki         & 10879         & 0.93362         & 0.00011         &          & 0.00818         &          &          \\

glv         & Manx         & 45473         & 0.997         & 0.00011         &          &          & 1.0         & 0.0         \\
\end{tabular}
}
\caption{Performance of \modelname on \corpusname test, \flores-200 and \udhr benchmarks (part 8)}\label{tab:appendix_glot_per_lang_8}
\end{table*}
\begin{table*}[h]
\small
\centering
\resizebox{0.57\textheight}{!}{
\begin{tabular}{llrrrrrrr}
& 
& 
& \multicolumn{2}{c}{\textbf{\corpusname}}
& \multicolumn{2}{c}{\textbf{\flores-200}} 
& \multicolumn{2}{c}{\textbf{\udhr}} 
\\
\cmidrule(lr){4-5} 
\cmidrule(lr){6-7}
\cmidrule(lr){8-9}
\textbf{iso639-3} &
\textbf{Language} &
\textbf{$|$Sentences$|$} &
\textbf{F1}$\uparrow$ & \textbf{FPR}$\downarrow$ &
\textbf{F1}$\uparrow$ &
\textbf{FPR}$\downarrow$ &
\textbf{F1}$\uparrow$ & 
\textbf{FPR}$\downarrow$ 
\\
\midrule
hmo         & Hiri Motu         & 1029005         & 0.97791         & 0.00113         &          &          &          &          \\

hmr         & Hmar         & 50021         & 0.99449         & 8e-05         &          &          &          &          \\

hne         & Chhattisgarhi         & 60771         & 0.95589         & 0.00017         & 0.90296         & 0.00256         &          & 0.00011         \\

hnj         & Hmong Njua         & 61134         & 0.99552         & 0.00022         &          &          &          &          \\

hnn         & Hanunoo         & 7931         & 0.9995         & 0.0         &          &          &          &          \\

hns         & Caribbean Hindustani         & 9791         & 0.99599         & 6e-05         &          &          & 0.88889         & 0.00033         \\
\end{tabular}
    }
    \caption{Performance of \modelname on \corpusname test, \flores-200 and \udhr benchmarks (part 9)}\label{tab:appendix_glot_per_lang_9}\end{table*}
\begin{table*}[h]
\small
\centering
\resizebox{0.57\textheight}{!}{
\begin{tabular}{llrrrrrrr}
& 
& 
& \multicolumn{2}{c}{\textbf{\corpusname}}
& \multicolumn{2}{c}{\textbf{\flores-200}} 
& \multicolumn{2}{c}{\textbf{\udhr}} 
\\
\cmidrule(lr){4-5} 
\cmidrule(lr){6-7}
\cmidrule(lr){8-9}
\textbf{iso639-3} &
\textbf{Language} &
\textbf{$|$Sentences$|$} &
\textbf{F1}$\uparrow$ & \textbf{FPR}$\downarrow$ &
\textbf{F1}$\uparrow$ &
\textbf{FPR}$\downarrow$ &
\textbf{F1}$\uparrow$ & 
\textbf{FPR}$\downarrow$ 
\\
\midrule
jac         & Popti'         & 8027         & 0.998         & 0.0         &          &          &          & 0.00011         \\

jae         & Yabem         & 7624         & 1.0         & 0.0         &          &          &          &          \\

jam         & Jamaican Creole English         & 23393         & 0.99299         & 0.00014         &          & 6e-05         &          &          \\

jav         & Javanese         & 233182         & 0.98066         & 0.00077         & 0.98346         & 0.00187         & 0.97581         & 0.00044         \\

jbo         & Lojban         & 16508         & 0.99347         & 6e-05         &          &          &          & 0.00099         \\

jbu         & Jukun Takum         & 7940         & 0.99599         & 6e-05         &          &          &          & 0.00044         \\
\end{tabular}
    }
    \caption{Performance of \modelname on \corpusname test, \flores-200 and \udhr benchmarks (part 10)}\label{tab:appendix_glot_per_lang_10}\end{table*}
\begin{table*}[h]
\small
\centering
\resizebox{0.57\textheight}{!}{
\begin{tabular}{llrrrrrrr}
& 
& 
& \multicolumn{2}{c}{\textbf{\corpusname}}
& \multicolumn{2}{c}{\textbf{\flores-200}} 
& \multicolumn{2}{c}{\textbf{\udhr}} 
\\
\cmidrule(lr){4-5} 
\cmidrule(lr){6-7}
\cmidrule(lr){8-9}
\textbf{iso639-3} &
\textbf{Language} &
\textbf{$|$Sentences$|$} &
\textbf{F1}$\uparrow$ & \textbf{FPR}$\downarrow$ &
\textbf{F1}$\uparrow$ &
\textbf{FPR}$\downarrow$ &
\textbf{F1}$\uparrow$ & 
\textbf{FPR}$\downarrow$ 
\\
\midrule
klt         & Nukna         & 752         & 0.99567         & 3e-05         &          &          &          &          \\

klv         & Maskelynes         & 7892         & 0.999         & 3e-05         &          &          &          &          \\

kma         & Konni         & 7936         & 0.999         & 6e-05         &          &          &          &          \\

kmb         & Kimbundu         & 385054         & 0.98706         & 0.0005         & 0.96321         & 0.00415         & 0.99194         & 0.00011         \\

kmg         & Kâte         & 6156         & 1.0         & 0.0         &          &          &          &          \\

kmh         & Kalam         & 15160         & 0.999         & 3e-05         &          &          &          &          \\
\end{tabular}
    }
    \caption{Performance of \modelname on \corpusname test, \flores-200 and \udhr benchmarks (part 11)}\label{tab:appendix_glot_per_lang_11}\end{table*}
\begin{table*}[h]
\small
\centering
\resizebox{0.56\textheight}{!}{
\begin{tabular}{llrrrrrrr}
& 
& 
& \multicolumn{2}{c}{\textbf{\corpusname}}
& \multicolumn{2}{c}{\textbf{\flores-200}} 
& \multicolumn{2}{c}{\textbf{\udhr}} 
\\
\cmidrule(lr){4-5} 
\cmidrule(lr){6-7}
\cmidrule(lr){8-9}
\textbf{iso639-3} &
\textbf{Language} &
\textbf{$|$Sentences$|$} &
\textbf{F1}$\uparrow$ & \textbf{FPR}$\downarrow$ &
\textbf{F1}$\uparrow$ &
\textbf{FPR}$\downarrow$ &
\textbf{F1}$\uparrow$ & 
\textbf{FPR}$\downarrow$ 
\\
\midrule
kvj         & Psikye         & 7908         & 1.0         & 0.0         &          &          &          &          \\

kvn         & Border Kuna         & 7926         & 1.0         & 0.0         &          &          &          &          \\

kwd         & Kwaio         & 7876         & 0.9995         & 3e-05         &          &          &          &          \\

kwf         & Kwara'ae         & 9038         & 0.9975         & 3e-05         &          &          &          &          \\

kwi         & Awa-Cuaiquer         & 7920         & 0.999         & 0.0         &          & 6e-05         & 0.84716         & 0.0         \\

kwj         & Kwanga         & 7936         & 1.0         & 0.0         &          &          &          &          \\
\end{tabular}
    }
    \caption{Performance of \modelname on \corpusname test, \flores-200 and \udhr benchmarks (part 12)}\label{tab:appendix_glot_per_lang_12}\end{table*}
\begin{table*}[h]
\small
\centering
\resizebox{0.57\textheight}{!}{
\begin{tabular}{llrrrrrrr}
& 
& 
& \multicolumn{2}{c}{\textbf{\corpusname}}
& \multicolumn{2}{c}{\textbf{\flores-200}} 
& \multicolumn{2}{c}{\textbf{\udhr}} 
\\
\cmidrule(lr){4-5} 
\cmidrule(lr){6-7}
\cmidrule(lr){8-9}
\textbf{iso639-3} &
\textbf{Language} &
\textbf{$|$Sentences$|$} &
\textbf{F1}$\uparrow$ & \textbf{FPR}$\downarrow$ &
\textbf{F1}$\uparrow$ &
\textbf{FPR}$\downarrow$ &
\textbf{F1}$\uparrow$ & 
\textbf{FPR}$\downarrow$ 
\\
\midrule
lol         & Mongo         & 7934         & 0.997         & 0.00011         &          &          &          &          \\

lom         & Loma (Liberia)         & 8918         & 0.96124         & 0.00014         &          &          &          &          \\

loq         & Lobala         & 3745         & 0.99573         & 3e-05         &          &          &          &          \\

lou         & Louisiana Creole         & 22         &          &          &          &          &          &          \\

loz         & Lozi         & 961751         & 0.99303         & 0.00033         &          & 0.00011         & 0.84892         & 0.0         \\

lsi         & Lashi         & 7897         & 0.9995         & 0.0         &          &          &          &          \\
\end{tabular}
    }
    \caption{Performance of \modelname on \corpusname test, \flores-200 and \udhr benchmarks (part 13)}\label{tab:appendix_glot_per_lang_13}\end{table*}
\begin{table*}[h]
\small
\centering
\resizebox{0.57\textheight}{!}{
\begin{tabular}{llrrrrrrr}
& 
& 
& \multicolumn{2}{c}{\textbf{\corpusname}}
& \multicolumn{2}{c}{\textbf{\flores-200}} 
& \multicolumn{2}{c}{\textbf{\udhr}} 
\\
\cmidrule(lr){4-5} 
\cmidrule(lr){6-7}
\cmidrule(lr){8-9}
\textbf{iso639-3} &
\textbf{Language} &
\textbf{$|$Sentences$|$} &
\textbf{F1}$\uparrow$ & \textbf{FPR}$\downarrow$ &
\textbf{F1}$\uparrow$ &
\textbf{FPR}$\downarrow$ &
\textbf{F1}$\uparrow$ & 
\textbf{FPR}$\downarrow$ 
\\
\midrule
mfh         & Matal         & 7949         & 1.0         & 0.0         &          &          &          &          \\

mfi         & Wandala         & 7930         & 1.0         & 0.0         &          &          &          & 0.00011         \\

mfk         & North Mofu         & 7937         & 0.99346         & 3e-05         &          &          &          &          \\

mfq         & Moba         & 9584         & 0.9965         & 8e-05         &          & 0.00045         & 0.88722         & 0.00164         \\

mfy         & Mayo         & 10521         & 0.99548         & 3e-05         &          &          &          & 0.00011         \\

mfz         & Mabaan         & 7874         & 1.0         & 0.0         &          &          &          &          \\
\end{tabular}
    }
    \caption{Performance of \modelname on \corpusname test, \flores-200 and \udhr benchmarks (part 14)}\label{tab:appendix_glot_per_lang_14}\end{table*}
\begin{table*}[h]
\small
\centering
\resizebox{0.57\textheight}{!}{
\begin{tabular}{llrrrrrrr}
& 
& 
& \multicolumn{2}{c}{\textbf{\corpusname}}
& \multicolumn{2}{c}{\textbf{\flores-200}} 
& \multicolumn{2}{c}{\textbf{\udhr}} 
\\
\cmidrule(lr){4-5} 
\cmidrule(lr){6-7}
\cmidrule(lr){8-9}
\textbf{iso639-3} &
\textbf{Language} &
\textbf{$|$Sentences$|$} &
\textbf{F1}$\uparrow$ & \textbf{FPR}$\downarrow$ &
\textbf{F1}$\uparrow$ &
\textbf{FPR}$\downarrow$ &
\textbf{F1}$\uparrow$ & 
\textbf{FPR}$\downarrow$ 
\\
\midrule
mqj         & Mamasa         & 8069         & 0.992         & 0.00022         &          &          &          & 0.00011         \\

mqy         & Manggarai         & 7953         & 0.9985         & 3e-05         &          &          &          &          \\

mrg         & Mising         & 6373         & 0.9995         & 0.0         &          &          &          & 0.00022         \\

mri         & Maori         & 79437         & 0.98948         & 0.00025         & 0.99901         & 6e-05         & 0.8227         & 0.00022         \\

mrj         & Western Mari         & 10083         & 0.99041         & 0.0         &          &          &          & 0.00011         \\

mrq         & North Marquesan         & 1007         & 0.96774         & 0.0         &          & 6e-05         &          & 0.00011         \\
\end{tabular}
    }
    \caption{Performance of \modelname on \corpusname test, \flores-200 and \udhr benchmarks (part 15)}\label{tab:appendix_glot_per_lang_15}\end{table*}
\begin{table*}[h]
\small
\centering
\resizebox{0.57\textheight}{!}{
\begin{tabular}{llrrrrrrr}
& 
& 
& \multicolumn{2}{c}{\textbf{\corpusname}}
& \multicolumn{2}{c}{\textbf{\flores-200}} 
& \multicolumn{2}{c}{\textbf{\udhr}} 
\\
\cmidrule(lr){4-5} 
\cmidrule(lr){6-7}
\cmidrule(lr){8-9}
\textbf{iso639-3} &
\textbf{Language} &
\textbf{$|$Sentences$|$} &
\textbf{F1}$\uparrow$ & \textbf{FPR}$\downarrow$ &
\textbf{F1}$\uparrow$ &
\textbf{FPR}$\downarrow$ &
\textbf{F1}$\uparrow$ & 
\textbf{FPR}$\downarrow$ 
\\
\midrule
nbc         & Chang Naga         & 7899         & 0.998         & 6e-05         &          &          &          &          \\

nbe         & Konyak Naga         & 7951         & 0.9995         & 0.0         &          &          &          &          \\

nbl         & South Ndebele         & 238555         & 0.9679         & 0.0008         &          &          & 0.0         & 0.00055         \\

nbq         & Nggem         & 3714         & 1.0         & 0.0         &          &          &          & 0.00033         \\

nbu         & Rongmei Naga         & 30767         & 0.99198         & 0.00017         &          &          &          &          \\

nca         & Iyo         & 7866         & 0.999         & 3e-05         &          &          &          &          \\
\end{tabular}
    }
    \caption{Performance of \modelname on \corpusname test, \flores-200 and \udhr benchmarks (part 16)}\label{tab:appendix_glot_per_lang_16}\end{table*}
\begin{table*}[h]
\small
\centering
\resizebox{0.57\textheight}{!}{
\begin{tabular}{llrrrrrrr}
& 
& 
& \multicolumn{2}{c}{\textbf{\corpusname}}
& \multicolumn{2}{c}{\textbf{\flores-200}} 
& \multicolumn{2}{c}{\textbf{\udhr}} 
\\
\cmidrule(lr){4-5} 
\cmidrule(lr){6-7}
\cmidrule(lr){8-9}
\textbf{iso639-3} &
\textbf{Language} &
\textbf{$|$Sentences$|$} &
\textbf{F1}$\uparrow$ & \textbf{FPR}$\downarrow$ &
\textbf{F1}$\uparrow$ &
\textbf{FPR}$\downarrow$ &
\textbf{F1}$\uparrow$ & 
\textbf{FPR}$\downarrow$ 
\\
\midrule
nno         & Norwegian Nynorsk         & 463877         & 0.98089         & 0.00036         & 0.98507         & 0.00045         & 0.95868         & 0.00033         \\

nnp         & Wancho Naga         & 7948         & 1.0         & 0.0         &          &          &          & 0.00011         \\

nnq         & Ngindo         & 7940         & 0.99649         & 6e-05         &          &          &          &          \\

nnw         & Southern Nuni         & 7874         & 1.0         & 0.0         &          &          &          &          \\

noa         & Woun Meu         & 15760         & 0.9995         & 0.0         &          &          &          & 0.00011         \\

nob         & Norwegian Bokmål         & 2890508         & 0.96898         & 0.00129         & 0.98185         & 0.00148         & 0.98462         & 0.00022         \\
\end{tabular}
    }
    \caption{Performance of \modelname on \corpusname test, \flores-200 and \udhr benchmarks (part 17)}\label{tab:appendix_glot_per_lang_17}\end{table*}
\begin{table*}[h]
\small
\centering
\resizebox{0.57\textheight}{!}{
\begin{tabular}{llrrrrrrr}
& 
& 
& \multicolumn{2}{c}{\textbf{\corpusname}}
& \multicolumn{2}{c}{\textbf{\flores-200}} 
& \multicolumn{2}{c}{\textbf{\udhr}} 
\\
\cmidrule(lr){4-5} 
\cmidrule(lr){6-7}
\cmidrule(lr){8-9}
\textbf{iso639-3} &
\textbf{Language} &
\textbf{$|$Sentences$|$} &
\textbf{F1}$\uparrow$ & \textbf{FPR}$\downarrow$ &
\textbf{F1}$\uparrow$ &
\textbf{FPR}$\downarrow$ &
\textbf{F1}$\uparrow$ & 
\textbf{FPR}$\downarrow$ 
\\
\midrule
ots         & Estado de México Otomi         & 13242         & 0.95737         & 0.00041         &          &          &          & 0.0046         \\

otw         & Ottawa         & 108         & 1.0         & 0.0         &          &          &          & 0.00044         \\

oym         & Wayampi         & 7693         & 0.9995         & 3e-05         &          &          &          &          \\

ozm         & Koonzime         & 7891         & 1.0         & 0.0         &          &          &          &          \\

pab         & Parecís         & 7936         & 0.9995         & 3e-05         &          &          &          &          \\

pad         & Paumarí         & 7771         & 1.0         & 0.0         &          &          &          &          \\
\end{tabular}
    }
    \caption{Performance of \modelname on \corpusname test, \flores-200 and \udhr benchmarks (part 18)}\label{tab:appendix_glot_per_lang_18}\end{table*}
\begin{table*}[h]
\small
\centering
\resizebox{0.57\textheight}{!}{
\begin{tabular}{llrrrrrrr}
& 
& 
& \multicolumn{2}{c}{\textbf{\corpusname}}
& \multicolumn{2}{c}{\textbf{\flores-200}} 
& \multicolumn{2}{c}{\textbf{\udhr}} 
\\
\cmidrule(lr){4-5} 
\cmidrule(lr){6-7}
\cmidrule(lr){8-9}
\textbf{iso639-3} &
\textbf{Language} &
\textbf{$|$Sentences$|$} &
\textbf{F1}$\uparrow$ & \textbf{FPR}$\downarrow$ &
\textbf{F1}$\uparrow$ &
\textbf{FPR}$\downarrow$ &
\textbf{F1}$\uparrow$ & 
\textbf{FPR}$\downarrow$ 
\\
\midrule
que         & Quechua         & 694846         & 0.97263         & 0.0014         &          &          & 0.9916         & 0.0         \\

quf         & Lambayeque Quechua         & 8072         & 0.98582         & 3e-05         &          & 6e-05         &          & 0.00394         \\

qug         & Chimborazo Highland Quichua         & 160594         & 0.84767         & 0.00787         &          &          & 0.0375         & 0.0         \\

quh         & South Bolivian Quechua         & 46618         & 0.81818         & 0.00861         &          & 0.00028         & 0.92174         & 0.00044         \\

qul         & North Bolivian Quechua         & 15852         & 0.77368         & 0.00223         &          & 6e-05         &          & 0.00011         \\

qup         & Southern Pastaza Quechua         & 7700         & 0.99701         & 0.00014         &          &          &          & 0.0093         \\
\end{tabular}
    }
    \caption{Performance of \modelname on \corpusname test, \flores-200 and \udhr benchmarks (part 19)}\label{tab:appendix_glot_per_lang_19}\end{table*}
\begin{table*}[h]
\small
\centering
\resizebox{0.57\textheight}{!}{
\begin{tabular}{llrrrrrrr}
& 
& 
& \multicolumn{2}{c}{\textbf{\corpusname}}
& \multicolumn{2}{c}{\textbf{\flores-200}} 
& \multicolumn{2}{c}{\textbf{\udhr}} 
\\
\cmidrule(lr){4-5} 
\cmidrule(lr){6-7}
\cmidrule(lr){8-9}
\textbf{iso639-3} &
\textbf{Language} &
\textbf{$|$Sentences$|$} &
\textbf{F1}$\uparrow$ & \textbf{FPR}$\downarrow$ &
\textbf{F1}$\uparrow$ &
\textbf{FPR}$\downarrow$ &
\textbf{F1}$\uparrow$ & 
\textbf{FPR}$\downarrow$ 
\\
\midrule
sby         & Soli         & 208         & 0.95833         & 0.0         &          &          &          & 0.00022         \\

sck         & Sadri         & 29         &          &          &          &          &          &          \\

scn         & Sicilian         & 62549         & 0.96888         & 0.00074         & 0.99802         & 6e-05         &          & 0.00263         \\

sco         & Scots         & 101211         & 0.97563         & 0.00025         &          & 0.00028         & 0.92683         & 0.00044         \\

sda         & Toraja-Sa'dan         & 8864         & 0.99198         & 0.00017         &          &          &          & 0.00011         \\

sdh         & Southern Kurdish         & 1048         & 0.93968         & 0.0         &          &          &          &          \\
\end{tabular}
    }
    \caption{Performance of \modelname on \corpusname test, \flores-200 and \udhr benchmarks (part 20)}\label{tab:appendix_glot_per_lang_20}\end{table*}
\begin{table*}[h]
\small
\centering
\resizebox{0.57\textheight}{!}{
\begin{tabular}{llrrrrrrr}
& 
& 
& \multicolumn{2}{c}{\textbf{\corpusname}}
& \multicolumn{2}{c}{\textbf{\flores-200}} 
& \multicolumn{2}{c}{\textbf{\udhr}} 
\\
\cmidrule(lr){4-5} 
\cmidrule(lr){6-7}
\cmidrule(lr){8-9}
\textbf{iso639-3} &
\textbf{Language} &
\textbf{$|$Sentences$|$} &
\textbf{F1}$\uparrow$ & \textbf{FPR}$\downarrow$ &
\textbf{F1}$\uparrow$ &
\textbf{FPR}$\downarrow$ &
\textbf{F1}$\uparrow$ & 
\textbf{FPR}$\downarrow$ 
\\
\midrule
stn         & Owa         & 7884         & 0.9985         & 6e-05         &          &          &          &          \\

stp         & Southeastern Tepehuan         & 7918         & 1.0         & 0.0         &          &          &          & 0.00022         \\

stq         & Saterfriesisch         & 10507         & 0.99147         & 0.00014         &          &          &          &          \\

sua         & Sulka         & 7558         & 0.999         & 3e-05         &          &          &          &          \\

suc         & Western Subanon         & 7896         & 0.999         & 0.0         &          &          &          &          \\

sue         & Suena         & 7882         & 0.998         & 8e-05         &          &          &          &          \\
\end{tabular}
    }
    \caption{Performance of \modelname on \corpusname test, \flores-200 and \udhr benchmarks (part 21)}\label{tab:appendix_glot_per_lang_21}\end{table*}
\begin{table*}[h]
\small
\centering
\resizebox{0.57\textheight}{!}{
\begin{tabular}{llrrrrrrr}
& 
& 
& \multicolumn{2}{c}{\textbf{\corpusname}}
& \multicolumn{2}{c}{\textbf{\flores-200}} 
& \multicolumn{2}{c}{\textbf{\udhr}} 
\\
\cmidrule(lr){4-5} 
\cmidrule(lr){6-7}
\cmidrule(lr){8-9}
\textbf{iso639-3} &
\textbf{Language} &
\textbf{$|$Sentences$|$} &
\textbf{F1}$\uparrow$ & \textbf{FPR}$\downarrow$ &
\textbf{F1}$\uparrow$ &
\textbf{FPR}$\downarrow$ &
\textbf{F1}$\uparrow$ & 
\textbf{FPR}$\downarrow$ 
\\
\midrule
tiv         & Tiv         & 481805         & 0.9965         & 0.00011         &          &          & 0.98551         & 0.00022         \\

tiy         & Tiruray         & 7896         & 1.0         & 0.0         &          &          &          &          \\

tke         & Takwane         & 10656         & 0.99498         & 6e-05         &          &          &          & 0.00011         \\

tkl         & Tokelau         & 629         & 0.91026         & 0.0         &          &          &          &          \\

tkr         & Tsakhur         & 912         & 0.98932         & 0.0         &          &          &          &          \\

tku         & Upper Necaxa Totonac         & 7936         & 0.99303         & 0.0003         &          &          &          &          \\
\end{tabular}
    }
    \caption{Performance of \modelname on \corpusname test, \flores-200 and \udhr benchmarks (part 22)}\label{tab:appendix_glot_per_lang_22}\end{table*}
\begin{table*}[h]
\small
\centering
\resizebox{0.56\textheight}{!}{
\begin{tabular}{llrrrrrrr}
& 
& 
& \multicolumn{2}{c}{\textbf{\corpusname}}
& \multicolumn{2}{c}{\textbf{\flores-200}} 
& \multicolumn{2}{c}{\textbf{\udhr}} 
\\
\cmidrule(lr){4-5} 
\cmidrule(lr){6-7}
\cmidrule(lr){8-9}
\textbf{iso639-3} &
\textbf{Language} &
\textbf{$|$Sentences$|$} &
\textbf{F1}$\uparrow$ & \textbf{FPR}$\downarrow$ &
\textbf{F1}$\uparrow$ &
\textbf{FPR}$\downarrow$ &
\textbf{F1}$\uparrow$ & 
\textbf{FPR}$\downarrow$ 
\\
\midrule
tzl         & Talossan         & 343         & 0.775         & 3e-05         &          &          &          &          \\

tzm         & Central Atlas Tamazight         & 8142         & 0.88564         & 0.00292         & 0.94421         & 0.0054         & 0.01754         & 0.00569         \\

tzo         & Tzotzil         & 521363         & 0.9985         & 8e-05         &          &          & 0.97479         & 0.00011         \\

ubr         & Ubir         & 9424         & 0.9975         & 8e-05         &          &          &          &          \\

ubu         & Umbu-Ungu         & 15261         & 0.9985         & 3e-05         &          &          &          &          \\

udm         & Udmurt         & 95670         & 0.99053         & 0.00036         &          &          &          & 0.00022         \\
\end{tabular}
    }
    \caption{Performance of \modelname on \corpusname test, \flores-200 and \udhr benchmarks (part 23)}\label{tab:appendix_glot_per_lang_23}\end{table*}
\begin{table*}[h]
\small
\centering
\resizebox{0.57\textheight}{!}{
\begin{tabular}{llrrrrrrr}
& 
& 
& \multicolumn{2}{c}{\textbf{\corpusname}}
& \multicolumn{2}{c}{\textbf{\flores-200}} 
& \multicolumn{2}{c}{\textbf{\udhr}} 
\\
\cmidrule(lr){4-5} 
\cmidrule(lr){6-7}
\cmidrule(lr){8-9}
\textbf{iso639-3} &
\textbf{Language} &
\textbf{$|$Sentences$|$} &
\textbf{F1}$\uparrow$ & \textbf{FPR}$\downarrow$ &
\textbf{F1}$\uparrow$ &
\textbf{FPR}$\downarrow$ &
\textbf{F1}$\uparrow$ & 
\textbf{FPR}$\downarrow$ 
\\
\midrule
wuv         & Wuvulu-Aua         & 7938         & 0.998         & 0.0         &          &          &          &          \\

wwa         & Waama         & 7901         & 0.9985         & 0.0         &          &          & 0.88636         & 0.0         \\

xal         & Kalmyk         & 8727         & 0.99147         & 0.00014         &          &          &          & 0.00022         \\

xav         & Xavánte         & 24284         & 0.78285         & 0.00803         &          &          &          & 0.00022         \\

xbi         & Kombio         & 4326         & 0.99924         & 0.0         &          &          &          &          \\

xbr         & Kambera         & 7547         & 0.999         & 0.0         &          &          &          &          \\
\end{tabular}
    }
    \caption{Performance of \modelname on \corpusname test, \flores-200 and \udhr benchmarks (part 24)}\label{tab:appendix_glot_per_lang_24}\end{table*}
\begin{table*}[h]
\small
\centering
\resizebox{0.57\textheight}{!}{
\begin{tabular}{llrrrrrrr}
& 
& 
& \multicolumn{2}{c}{\textbf{\corpusname}}
& \multicolumn{2}{c}{\textbf{\flores-200}} 
& \multicolumn{2}{c}{\textbf{\udhr}} 
\\
\cmidrule(lr){4-5} 
\cmidrule(lr){6-7}
\cmidrule(lr){8-9}
\textbf{iso639-3} &
\textbf{Language} &
\textbf{$|$Sentences$|$} &
\textbf{F1}$\uparrow$ & \textbf{FPR}$\downarrow$ &
\textbf{F1}$\uparrow$ &
\textbf{FPR}$\downarrow$ &
\textbf{F1}$\uparrow$ & 
\textbf{FPR}$\downarrow$ 
\\
\midrule
zdj         & Ngazidja Comorian         & 2097         & 0.97484         & 6e-05         &          & 0.00011         & 0.78125         & 0.00208         \\

zea         & Zeeuws         & 10012         & 0.91037         & 0.00025         &          & 0.00045         &          &          \\

zgh         & Standard Moroccan Tamazight         & 3285         & 0.93333         & 0.00039         &          & 0.00068         & 0.17647         & 0.0         \\

zho         & Chinese         & 4212269         & 0.998         & 0.0         & 1.0         & 0.0         &          &          \\

zia         & Zia         & 7893         & 0.99749         & 0.0         &          &          &          &          \\

ziw         & Zigula         & 7957         & 0.97857         & 0.00069         &          &          &          &          \\
\end{tabular}
    }
    \caption{Performance of \modelname on \corpusname test, \flores-200 and \udhr benchmarks (part 25)}\label{tab:appendix_glot_per_lang_25}\end{table*}

\section{Calibration}\seclabel{calibration}

As stated in \secref{reqs}, an LID model should provide a calibrated confidence measure in addition to its prediction. Reliability diagrams illustrate model calibration~\citep{degroot1983comparison, niculescu2005predicting}. These diagrams use expected sample accuracy as a function of confidence. If the model is perfectly calibrated, then the diagram plots the identity function.

\begin{figure}[ht]
    \centering
    \includegraphics[width=0.49\textwidth]{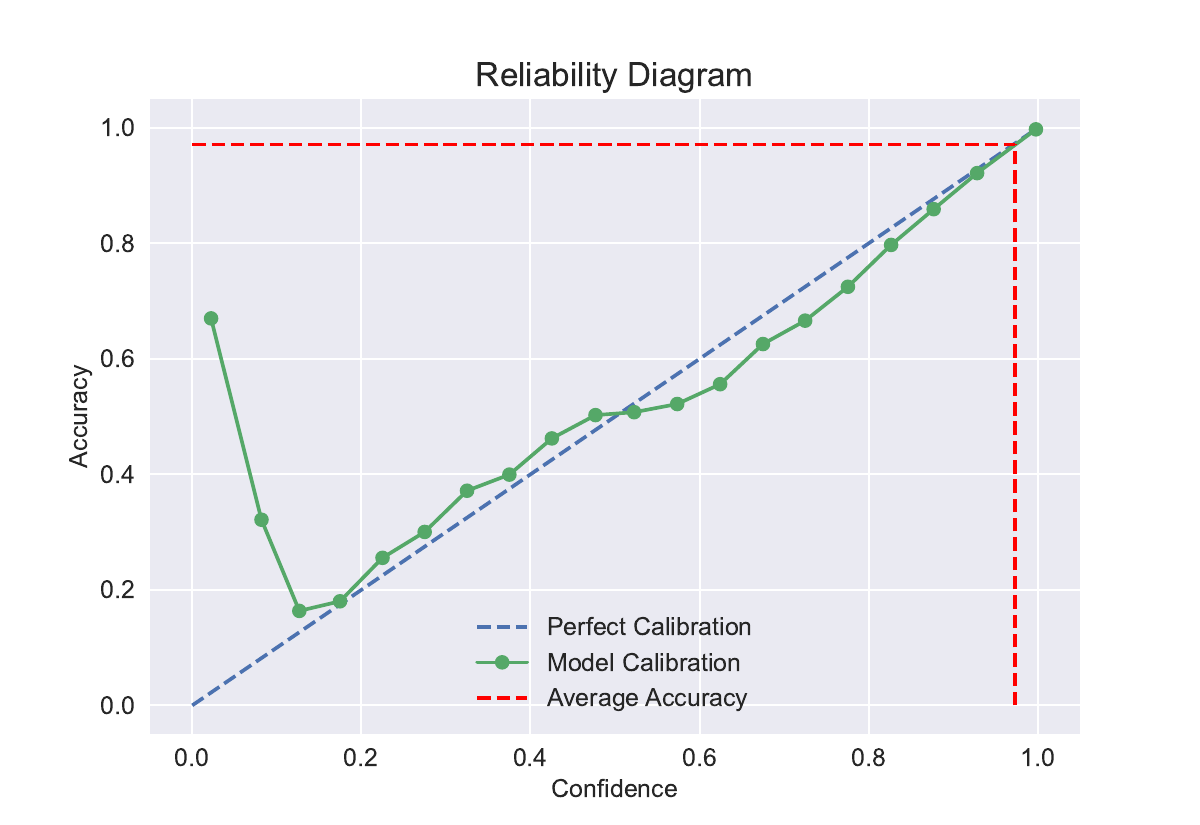}
    \caption{Reliability diagram for \modelname on \corpusname test}
    \label{fig:reliability-test}
\end{figure}

We provide the reliability diagram for \modelname on \corpusname test in \figref{reliability-test}.
For \corpusname test, the plot is nearly close to the identity function. However, for some of the low confidence scores, it's not calibrated. This mostly happens because we included so many languages in our models, and some of these languages are very similar to each other or have small training sizes.

\end{document}